%% file: main.tex
\documentclass{article}

\usepackage[final]{neurips_2026}

\usepackage[utf8]{inputenc} 
\usepackage[T1]{fontenc}    
\usepackage{url}            
\usepackage{booktabs}       
\usepackage{amsfonts}       
\usepackage{nicefrac}       
\usepackage{microtype}      
\usepackage{soul,xcolor}    
\usepackage{xspace}
\usepackage{graphicx}
\usepackage{amsmath}
\usepackage{bm}
\usepackage{tcolorbox}
\usepackage{algorithm}
\usepackage{setspace}
\usepackage{algpseudocode}
\usepackage{arydshln}
\usepackage{amssymb}
\usepackage{bbm, dsfont}
\usepackage{mathtools}
\usepackage{amsthm}
\usepackage[normalem]{ulem}
\usepackage{colortbl}
\usepackage{multirow}
\usepackage{pifont}
\usepackage{eqparbox}
\usepackage{makecell}
\usepackage{threeparttable}
\usepackage{adjustbox}
\usepackage{titletoc}
\usepackage{tocloft}
\usepackage{enumitem}
\usepackage{caption}
\usepackage{wrapfig} 

\definecolor{darkgreen}{RGB}{0,128,0}
\definecolor{lightblue}{rgb}{0.9,0.95,1.0}
\definecolor{lightgray}{RGB}{240, 243, 246}

\newcommand{\ours}{\textsc{StreamGaze}\xspace}

\definecolor{pastColor}{HTML}{E7F2DC}
\definecolor{presentColor}{HTML}{FBF5DC}
\definecolor{proactiveColor}{HTML}{FBE4E7}

\input{preamble}

\definecolor{cvprblue}{rgb}{0.21,0.49,0.74}
\usepackage[pagebackref,breaklinks,colorlinks,allcolors=cvprblue]{hyperref}
\usepackage[capitalize,noabbrev]{cleveref}

\newcommand{\ie}{\emph{i.e.}\xspace}
\newcommand{\eg}{\emph{e.g.}\xspace}

\title{\ours{}: Gaze-Guided Temporal Reasoning \\ and Proactive Understanding in Streaming Videos}

\author{%
  Daeun Lee$^{1}$ \quad
  Subhojyoti Mukherjee$^{2}$ \quad
  Branislav Kveton$^{2}$ \quad
  Ryan A. Rossi$^{2}$ \\
  \textbf{Viet Dac Lai$^{2}$} \quad
  \textbf{Seunghyun Yoon$^{2}$} \quad
  \textbf{Trung Bui$^{2}$} \quad
  \textbf{Franck Dernoncourt$^{2}$} \quad
  \textbf{Mohit Bansal$^{1}$}\\
  $^{1}$University of North Carolina, Chapel Hill \quad
  $^{2}$Adobe Research\\
}

\begin{document}

\maketitle


\input{sec/0_abstract}
\input{sec/1_intro}
\input{sec/2_related}
\input{sec/3_method}
\input{sec/4_exps}
\input{sec/5_conclusion}

\bibliographystyle{plainnat}
\bibliography{main}

\appendix
\input{sec/supple}


\end{document}

%% file: preamble.tex
\usepackage[textsize=tiny]{todonotes}



%% file: sec/0_abstract.tex
\begin{abstract}
Streaming video understanding requires models not only to process temporally incoming frames, but also to anticipate user intent 
for realistic applications such as Augmented Reality (AR) glasses. 
While prior streaming benchmarks evaluate temporal reasoning, none measure whether Multimodal Large Language Models (MLLMs) can interpret or leverage human gaze signals within a streaming setting. 
To fill this gap, we introduce \ours{}, the first benchmark designed to evaluate how effectively MLLMs utilize gaze for temporal and proactive reasoning in streaming videos. 
\ours{} introduces gaze-guided past, present, and proactive tasks
that comprehensively assess streaming video understanding.
Past and present tasks require immediate responses based on
observations available at query time, whereas proactive tasks
require models to continuously process incoming frames and
respond at the appropriate future moment.
To build \ours{}, we develop a gaze–video Question Answering (QA) generation pipeline that aligns egocentric videos with raw gaze trajectories through fixation extraction, region-specific visual prompting, and scanpath construction. This pipeline produces spatio-temporally grounded QA pairs that reflect human perceptual dynamics. 
Across all \ours{} tasks, we observe substantial performance gaps between state-of-the-art MLLMs and human performance, highlighting key limitations in gaze-based temporal reasoning, intent modeling, and proactive prediction.  
We further provide detailed analyses of gaze prompting strategies, reasoning behaviors, and task-specific failure modes, offering insights into current limitations and directions for future research. 
All data and code are publicly available to support research in gaze-guided streaming video understanding.
\end{abstract}

%% file: sec/1_intro.tex
\section{Introduction}

Recent advances in Video Large Language Models (VideoLLMs)~\citep{lin2023video,damonlpsg2023videollama,wang2025internvl3} 
have significantly improved multimodal understanding of dynamic visual environments. A growing line of work now investigates \textit{streaming video understanding}~\citep{streambench,chen2024videollm,fu2025vispeak,zhang2024flashvstreammemorybasedrealtimeunderstanding},
where models must process temporally incoming frames and respond in real time without access to future context. Such capability is essential for real-world applications such as robotics, embodied agents, and AR-glass assistants, where perception and decision-making must occur continuously as events unfold.

\begin{figure}[t]
\centering
\vspace{-1.5em} 
\includegraphics[width=0.99\linewidth]{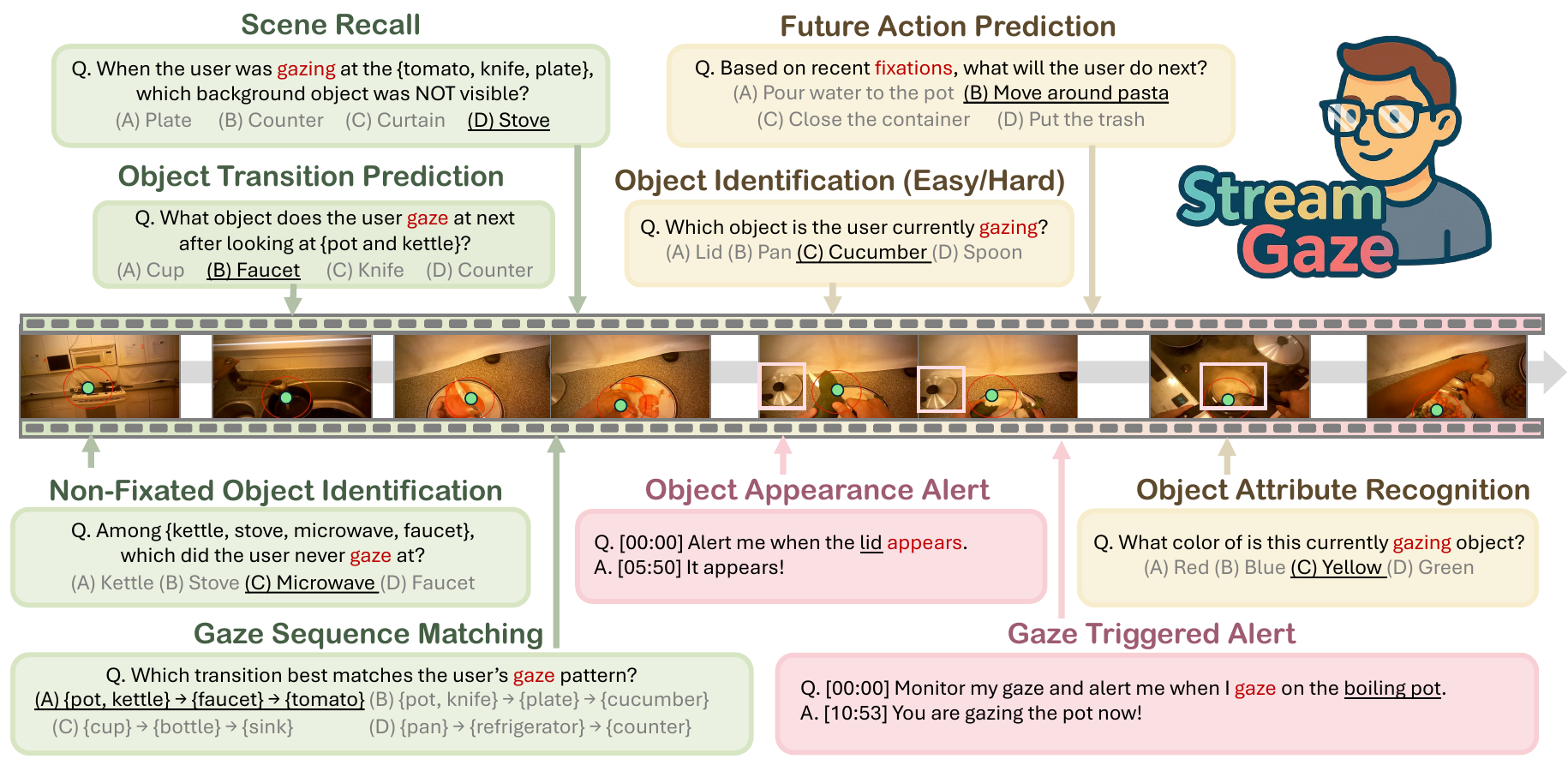}
\caption{\small\textbf{\ours{}'s task taxonomy.} We introduce \ours{}, the first benchmark designed to evaluate how effectively MLLMs use gaze for temporal and proactive reasoning in streaming videos. We introduce gaze-guided \colorbox{pastColor}{past}, \colorbox{presentColor}{present}, and \colorbox{proactiveColor}{proactive} streaming video understanding tasks.}
\label{fig:tasks}
\vspace{-.25in}
\end{figure}

Benchmarks play a critical role in tracking progress in this rapidly developing domain. While several streaming benchmarks exist~\citep{chen2024videollm,fu2025vispeak,wang2025omnimmi,yang2025svbench,zhang2025proactive}, they capture only some challenges faced by realistic streaming agents.
As summarized in \Cref{tab:benchmark_comparision}, some benchmarks~\citep{yang2025svbench,streambench} focus primarily on past or present recognition, leaving out proactive understanding, the ability to anticipate future events and user intents. 
Moreover, existing benchmarks rarely incorporate the human perceptual signals that actually drive real-world decision-making, even when they rely on egocentric video sources and implicitly assume AR-glasses usage scenarios.
Among these perceptual cues, \textit{eye gaze} is the most direct and reliable indicator of where a user is attending, what they are processing, and what they are likely to do next~\citep{gazean1,gazean2,ilaslan2023gazevqa,egtea}.
Omitting gaze removes the key perceptual signal humans use to filter relevant information, plan actions, and anticipate what will happen next. This leads to benchmark evaluations that diverge from how humans perceive and reason about dynamic environments.

However, incorporating gaze into video understanding has been challenging~\citep{peng2025eye,ilaslan2023gazevqa}. 
Unlike conventional QA that relies solely on visible content, gaze-guided QA requires interpreting \textit{dynamic gaze trajectories}, modeling how attention shifts over time, and grounding these signals within a moving egocentric viewpoint. 
Raw gaze streams are noisy and contain fast, unstable fluctuations~\citep{fov1,fov2,fov3}, while egocentric videos continuously shake and rotate with user motion, making even \textit{"what the user is looking at"} a non-trivial spatio-temporal grounding problem.
Solving these tasks requires models to interpret gaze online and reason \textit{causally} over streaming video: past and present tasks require immediate responses, while proactive tasks process the incoming stream until a future response.
Together, these factors make the automatic construction of temporally grounded QA pairs highly challenging. 

To close this gap, we introduce \ours{}, the first benchmark for \textbf{gaze-guided streaming video understanding}. 
As shown in \Cref{fig:tasks}, \ours{} provides a unified suite of gaze-conditioned tasks spanning \textit{past}, \textit{present}, and \textit{proactive} reasoning, enabling comprehensive evaluation of how well MLLMs use gaze for temporal understanding and future prediction. 
Solving these tasks requires models to interpret gaze online, follow shifting attention over time, and infer intent using only past and currently observed frames, mirroring the constraints faced by real streaming agents.

To construct \ours{}, we develop a semi-automatic gaze-guided data generation pipeline that aligns gaze trajectories with egocentric videos and converts them into temporally grounded QA pairs.
We first detect stable gaze moments (\ie, fixations) throughout each video and extract objects within the corresponding gaze regions using region-specific visual prompting.
Next, we model the temporal dynamics of gaze by constructing \textit{scanpaths}, defined as ordered sequences of fixated objects over time. 
These scanpaths capture how user attention shifts across objects and are used to identify relevant and irrelevant objects when forming gaze-conditioned queries. 
Building on these signals, we automatically generate past, present, and proactive streaming tasks using an LLM, followed by human verification for quality assurance. 

Across all \ours{} tasks, we observe substantial performance gaps between state-of-the-art MLLMs (e.g., Claude Opus 4.8~\citep{anthropic2025claude4}, Gemini 3.1 Pro~\citep{comanici2025gemini}, GPT-4o~\citep{openai2024gpt4technicalreport}, InternVL-3.5~\citep{wang2025internvl3}) and both human performance and specialized streaming video understanding models. 
In particular, current MLLMs struggle to leverage gaze signals for temporal reasoning and proactive inference, revealing fundamental limitations in gaze-conditioned understanding.
We further conduct detailed analyses of gaze-prompting strategies, gaze-based reasoning behaviors, and task-specific proactive responses on \ours{}.
Together, these evaluations position \ours{} as the first comprehensive testbed for assessing gaze-driven causal reasoning and proactive prediction in streaming video scenarios.

\input{table/comparision}

\noindent Our contributions are threefold: 
\begin{itemize}
    \item 
    We propose the first gaze-guided data construction pipeline that integrates gaze trajectories with egocentric video to produce spatio-temporally aligned, gaze-guided QA pairs.
    Our pipeline models full \textit{scanpath} dynamics, tracking how attention evolves over time and how new objects enter the FOV through ego-motion, enabling truly streaming, temporally grounded supervision that static gaze–based pipelines cannot provide.

    \item 
    We introduce \ours{}, the first benchmark specifically designed for streaming gaze-guided video understanding, comprising 8521 QA pairs across 10 past, present, and proactive tasks. This represents a larger and more diverse evaluation suite compared to existing streaming QA benchmarks.

    \item 
    We evaluate state-of-the-art MLLMs on \ours{}, uncovering substantial and consistent gaps relative to human performance. Our in-depth analyses of gaze-prompting strategies, reasoning patterns, and task-specific behaviors reveal that current MLLMs struggle to interpret raw gaze signals, overly rely on frame-local visual cues, and fail to generalize across different temporal reasoning requirements. These findings provide concrete design guidance for future gaze-aware streaming models.
\end{itemize}

%% file: table/comparision.tex
\begin{figure*}[t]
    \centering
    \begin{minipage}[t]{0.63\textwidth}  
    \centering
    \footnotesize
    
    \captionof{table}{
    \small\textbf{Comparison of streaming video understanding benchmarks.}
    MC: Multiple-choice; OE: Open-ended; All-time: Coverage of past, present, and future tasks; Proac.: Proactive; Ego.: Egocentric videos.
    \textcolor{darkgreen}{\ding{51}} indicates full support, while
    \textcolor{darkgreen}{\large $\triangle$} indicates partial support.
    }
    \label{tab:benchmark_comparision}
    
    \resizebox{\textwidth}{!}{   
    \begin{tabular}{l|cccccccc}
    \toprule
    \multicolumn{1}{c|}{Dataset} 
    & \makecell{Avg\\video (s)}
    & \makecell{QA\\pairs}
    & \makecell{QA\\type}
    & \makecell{Annotation}
    & \makecell{Proac.}
    & \makecell{Gaze}
    & \makecell{All-\\time}
    & \makecell{Ego.} \\ 
    \midrule
    \rowcolor{lightgray} \multicolumn{9}{l}{\textit{Gaze-based QA benchmarks}}\\
    
    GazeVQA~\cite{ilaslan2023gazevqa} 
    & \multicolumn{1}{c}{150} & 25040 & MC & Human 
    & \textcolor{red}{\ding{56}} & \textcolor{darkgreen}{\ding{51}}
    & \textcolor{red}{\ding{56}} & \textcolor{darkgreen}{\ding{51}} \\
    
    EgoGazeVQA~\cite{peng2025eye} 
    & \multicolumn{1}{c}{52} & 1757 & MC & Auto
    & \textcolor{red}{\ding{56}} & \textcolor{darkgreen}{\ding{51}}
    & \textcolor{red}{\ding{56}} & \textcolor{darkgreen}{\ding{51}} \\
    
    \midrule
    \rowcolor{lightgray} \multicolumn{9}{l}{\textit{Streaming QA benchmarks}}\\
    
    StreamingBench~\cite{lin2024streamingbench} 
    & \multicolumn{1}{c}{268} & 4500 & MC+OE & Auto \& Human
    & \textcolor{darkgreen}{\ding{51}} & \textcolor{red}{\ding{56}}
    & \textcolor{darkgreen}{\ding{51}} & \textcolor{red}{\ding{56}} \\
    
    SVBench~\cite{yang2025svbench} 
    & \multicolumn{1}{c}{147} & 7374 & OE & Auto \& Human
    & \textcolor{red}{\ding{56}} & \textcolor{red}{\ding{56}}
    & \textcolor{red}{\ding{56}} & \textcolor{darkgreen}{\large $\triangle$} \\
    
    OVO-Bench~\cite{niu2025ovo} 
    & \multicolumn{1}{c}{428} & 2814 & MC+OE & Auto \& Human
    & \textcolor{darkgreen}{\ding{51}} & \textcolor{red}{\ding{56}}
    & \textcolor{darkgreen}{\ding{51}} & \textcolor{darkgreen}{\large $\triangle$} \\
    
    OmniMMI~\cite{wang2025omnimmi} 
    & \multicolumn{1}{c}{324} & 2290 & OE & Auto \& Human
    & \textcolor{darkgreen}{\ding{51}} & \textcolor{red}{\ding{56}}
    & \textcolor{red}{\ding{56}} & \textcolor{darkgreen}{\large $\triangle$} \\
    
    ProAssist~\cite{zhang2025proactive} 
    & \multicolumn{1}{c}{710} & 1020 & OE & Auto
    & \textcolor{darkgreen}{\ding{51}} & \textcolor{red}{\ding{56}}
    & \textcolor{red}{\ding{56}} & \textcolor{darkgreen}{\ding{51}} \\
    
    StreamBench~\cite{streambench} 
    & \multicolumn{1}{c}{270} & 1800 & OE & Auto \& Human
    & \textcolor{red}{\ding{56}} & \textcolor{red}{\ding{56}}
    & \textcolor{red}{\ding{56}} & \textcolor{darkgreen}{\ding{51}} \\
    
    Vispeak-Bench~\cite{fu2025vispeak} 
    & \multicolumn{1}{c}{21} & 1000 & OE & Auto \& Human
    & \textcolor{darkgreen}{\ding{51}} & \textcolor{red}{\ding{56}}
    & \textcolor{red}{\ding{56}} & \textcolor{red}{\ding{56}} \\
    
    \midrule
    \rowcolor{lightblue}
    \ours{} (Ours) 
    & \multicolumn{1}{c}{815} & 8521 & MC+OE & Auto \& Human
    & \textcolor{darkgreen}{\ding{51}} & \textcolor{darkgreen}{\ding{51}}
    & \textcolor{darkgreen}{\ding{51}} & \textcolor{darkgreen}{\ding{51}} \\
    
    \bottomrule
    \end{tabular}
    }
    \end{minipage}%
    \hfill
    \begin{minipage}[t]{0.33\textwidth}
    \vspace{0pt}
    \centering
    \includegraphics[width=\textwidth]{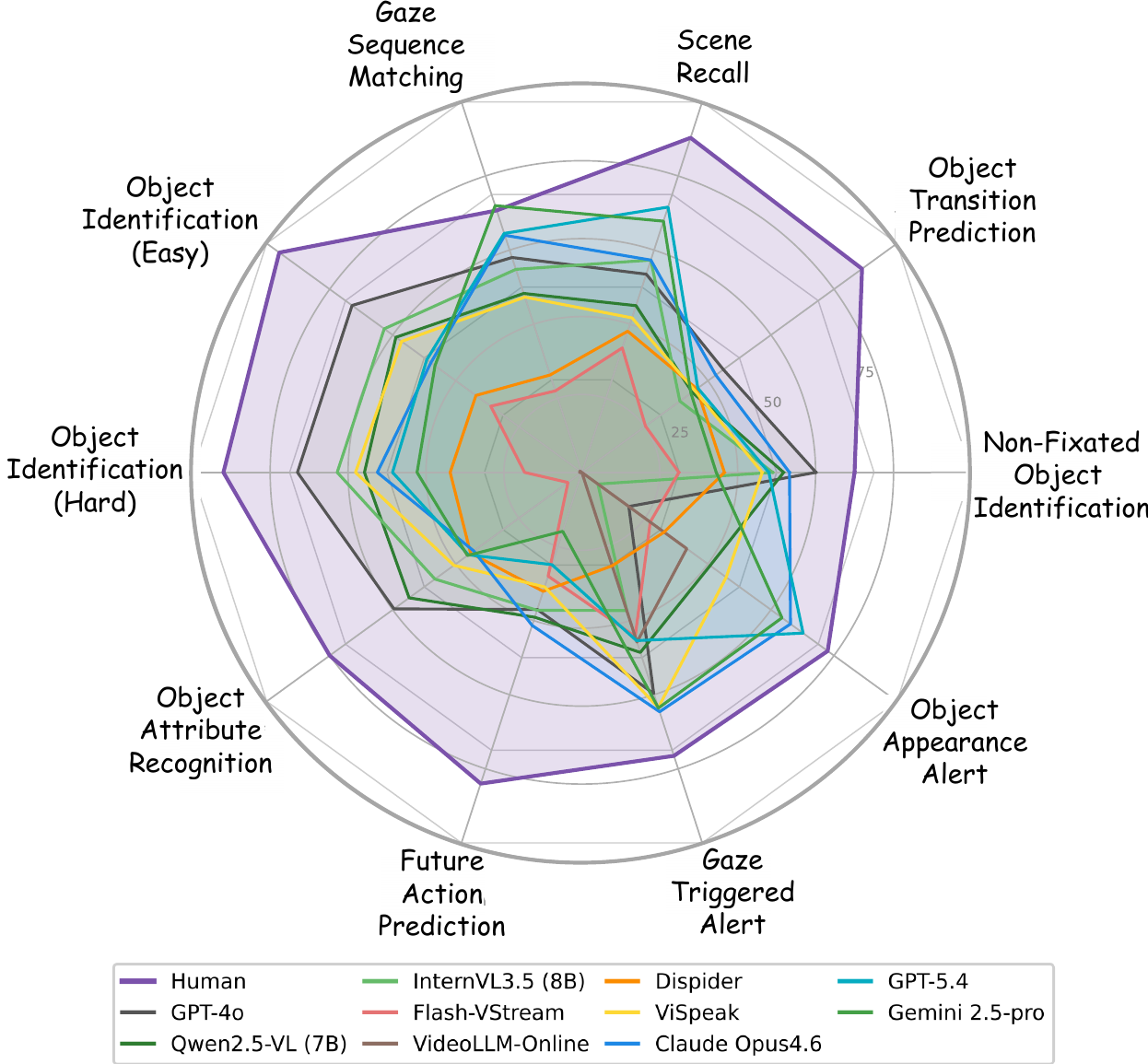}
    \captionof{figure}{\small\textbf{MLLMs performance across \ours{} tasks.} 
    }
    \label{fig:performance_radar}
    \end{minipage}
    \vspace{-0.20in}
    \end{figure*}

%% file: sec/2_related.tex
\section{Related Work}

\textbf{Streaming QA benchmarks.} 
Streaming video understanding requires models to interpret and respond to temporally incoming frames without access to future context.
StreamingBench~\citep{lin2024streamingbench} and OVO-Bench~\citep{niu2025ovo} benchmark online video understanding across past, present, and future tasks, providing a comprehensive evaluation of temporal reasoning capabilities.
Meanwhile, OmniMMI~\citep{wang2025omnimmi}, ProAssist~\citep{zhang2025proactive}, and ViSpeak-Bench~\citep{fu2025vispeak} extend this paradigm to interactive dialogue settings, assessing how effectively MLLMs perceive user intents and sustain coherent interactions in streaming contexts.
In contrast, our proposed \ours{} introduces \textit{a novel dimension by integrating eye-gaze dynamics into the streaming setting}, 
enabling systematic evaluation of how models leverage human gaze for temporal reasoning, intent inference, and proactive prediction.

\textbf{Gaze-based QA benchmarks.} 
GazeVQA~\citep{ilaslan2023gazevqa} introduces the first gaze-based VQA dataset designed for collaborative interactions in assembly processes.
EgoGazeVQA~\citep{peng2025eye} focuses primarily on intent understanding in short egocentric videos, but relies only on frame-wise gaze signals without extracting fixations or establishing spatio-temporal grounding for QA generation.
None of these works address the online setting that encompasses past, present, and proactive tasks, an essential component for real-world gaze-guided applications.
\ours{} is the first benchmark to integrate gaze dynamics into the \textit{streaming} setting, enabling systematic evaluation of how well MLLMs perform temporal reasoning and proactively align with human attention over time.

%% file: sec/3_method.tex
\section{Gaze-Guided Streaming Data Construction}\label{sec:pipeline}

\begin{figure*}[t]
    \centering
    {
    \includegraphics[width=\textwidth]{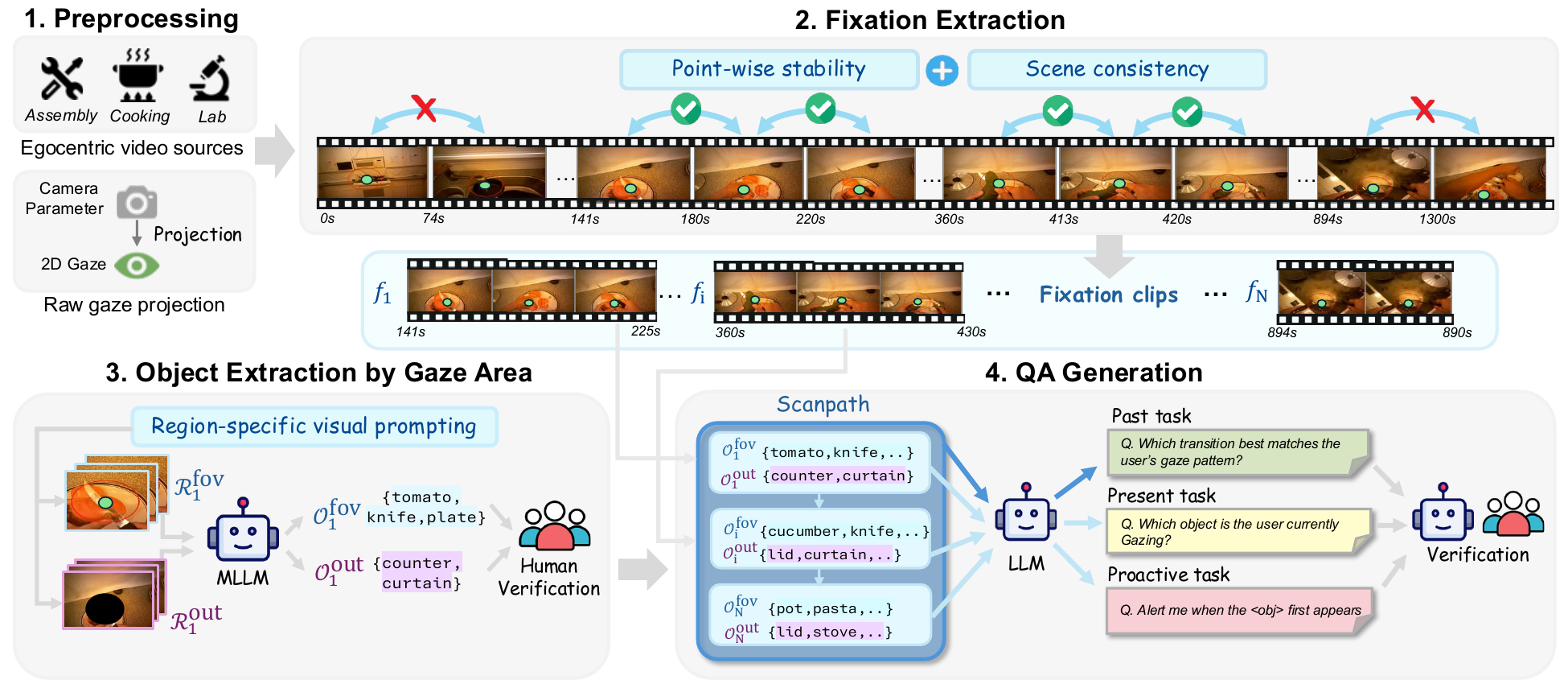}}
    \caption{
    \small\textbf{Gaze-guided streaming data construction pipeline.}
    Given egocentric video and raw gaze projections (\Cref{sec:approach-preproc}), we first extract fixation moments across the entire video (\Cref{sec:approach-fixation-extraction}).
    Next, we divide each frame into FOV and out-of-FOV regions and extract objects within the gaze area (\Cref{sec:approach-object-extraction}).
    Finally, we construct scanpaths and generate streaming QA pairs (\Cref{sec:benchmark}).
    }
    \label{fig:main_pipeline}
    \vspace{-0.1in}
\end{figure*}

We design a novel data construction pipeline to build gaze-guided streaming video understanding tasks. 
In contrast to EgoGazeVQA~\citep{peng2025eye}, which processes short clips with static per-frame gaze points, our pipeline extracts the full scanpath, a sequence of fixations that captures how attention shifts over time and how objects enter the user's FOV (Field of View) through ego-motion.
This enables streaming and temporally grounded data construction beyond static-gaze settings.

We first preprocess raw gaze trajectories by projecting them using camera parameters (\Cref{sec:approach-preproc}) and extract stable fixation moments that capture meaningful, attention-driven events (\Cref{sec:approach-fixation-extraction}).
Next, we divide each frame into FOV and out-of-FOV regions and extract the corresponding objects.
Finally, we construct scanpaths and conduct human verification—not only to ensure quality and consistency, but also to confirm that our pipeline’s data construction closely correlates with human annotations (\Cref{sec:approach-object-extraction}). 
Only human-verified scanpaths and object annotations are retained for the final benchmark, ensuring annotation reliability and quality (\Cref{sec:approach-human}).

\subsection{Preprocessing}\label{sec:approach-preproc}

We use three public egocentric video datasets spanning diverse domains: EGTEA+ (\textit{cooking})~\citep{egtea}, EgoExoLearn (\textit{cooking}, \textit{lab})~\citep{huang2024egoexolearn}, and HoloAssist (\textit{assembly})~\citep{wang2023holoassist}.
Each video is represented as a sequence of frames,
$\mathcal{V} = \{I_t\}_{t=1}^{T}$, where $T$ is the number of frames and $I_t$ is the frame at time $t$. 
To obtain the gaze trajectory
$\mathcal{G} = \{(x_t, y_t)\}_{t=1}^{T}$,
where $(x_t, y_t)$ denotes the gaze coordinate on the image plane corresponding to frame $I_t$,
we project raw gaze in world coordinates onto the 2D image plane using officially provided camera parameters. 
For datasets that directly provide 2D gaze coordinates (e.g.,~\citep{egtea,huang2024egoexolearn}), we use them.

\subsection{Fixation Extraction}
\label{sec:approach-fixation-extraction}
Based on the unified gaze trajectory $\mathcal{G}$ obtained in
\Cref{sec:approach-preproc},
we identify \textit{query moments} within $\mathcal{V}$,
corresponding to periods of meaningful user attention
for streaming video QA tasks.
We focus on \textit{fixation moments}, during which gaze remains
relatively stable within a localized region, as distinguished
from rapid eye movements (\ie, saccades)
~\citep{salvucci2000identifying,inadumi2024gaze,ilaslan2023gazevqa}.
These fixation moments provide reliable spatial and temporal
anchors for object-centric QA generation, enabling consistent,
reproducible, and objectively verifiable annotations.
While this design does not explicitly capture dynamic gaze
behaviors such as smooth pursuit and the vestibulo-ocular reflex,
it avoids reliance on additional signals such as
inertial measurement unit (IMU) data or head pose,
facilitating application to diverse public egocentric datasets.
Incorporating dynamic gaze behaviors remains an important
direction for future work.

From a raw gaze trajectory $\mathcal{G}$, we identify fixation
intervals by grouping temporally consecutive gaze observations
that exhibit spatial stability, while allowing brief gaze
interruptions of up to 200 ms to avoid premature segmentation.
Each fixation is characterized by a spatial centroid
$(\bar{x}_i, \bar{y}_i)$ and a temporal span
$[t_i^s, t_i^e] \subseteq [1,T]$,
where $t_i^s$ and $t_i^e$ denote the start and end timestamps.
If $N$ fixations are detected, the resulting set is written as
\begin{equation}
\mathcal{F} =
\{ f_i = (\bar{x}_i, \bar{y}_i, t_i^s, t_i^e) \}_{i=1}^{N}.
\label{eq:fixation}
\end{equation}
The spatial centroid is obtained by averaging all gaze points
within each interval.
We validate candidate fixations using two criteria:
\textit{(i) point-wise stability} and
\textit{(ii) scene consistency}.
Further implementation details are provided in
\Cref{sec:appendix:fixation}.

\textbf{Point-wise stability.}
We evaluate the spatial and temporal stability of each
candidate fixation interval.
To assess spatial stability, we compute the centroid
$(\bar{x}_i, \bar{y}_i)$ as the average of gaze points within
the interval and require that each point lies within a
maximum spatial radius:
\begin{equation}
\|(x_t, y_t) - (\bar{x}_i,\bar{y}_i)\|_2
\leq r_{\text{thresh}},
\quad \forall t \in [t_i^s, t_i^e],
\end{equation}
where $r_{\text{thresh}}$ denotes the maximum allowable spatial
dispersion around the centroid.
To ensure consistency across datasets with varying resolutions,
we normalize $r_{\text{thresh}}$ by the frame width.
Brief deviations lasting at most 200 ms are tolerated
to accommodate transient gaze interruptions.

We further enforce temporal stability by requiring that
the interval duration exceeds a minimum threshold:
\begin{equation}
t_i^e - t_i^s \geq \tau_{\text{dur}},
\end{equation}
where $\tau_{\text{dur}}$ is the minimum duration for a valid fixation.

\textbf{Scene consistency.}
Even when a candidate fixation satisfies spatial and temporal
stability, abrupt scene changes may occur due to camera motion
or cuts.
To ensure visual continuity, we uniformly sample frames from
the corresponding video segment
$\mathcal{V}_i = \{I_t\}_{t=t_i^s}^{t_i^e}$
and compute normalized Hue--Saturation histograms
$H_t$~\citep{radwan2012histogram}.
We then measure the minimum Pearson correlation between
consecutive sampled frames:
\begin{equation}
S_{\min} =
\min_{j \in \{1,\ldots,K-1\}}
\rho(H_{t_j}, H_{t_{j+1}}),
\end{equation}
where $K$ is the number of uniformly sampled frames,
$t_j$ denotes the timestamp of the $j$-th sampled frame,
and $\rho(\cdot)$ denotes the Pearson correlation between
normalized histograms.
A fixation is retained only if
$S_{\min} \geq \tau_{\text{scene}}$,
ensuring that visually discontinuous intervals are discarded.

\subsection{Object Extraction by Gaze Area} \label{sec:approach-object-extraction}
Based on each fixation $f_i$, 
we extract objects according to their locations within or outside the FOV from fixation video clip $\mathcal{V}_i$. 
Our objective is to distinguish objects that lie \textit{inside} the user’s FOV from those \textit{outside} it, enabling controllable task difficulty and gaze-grounded QA generation.

\textbf{Definition of FOV and out-of-FOV regions.}
Motivated by classic 
eye-tracking literature, which models the foveal and parafoveal regions as circular areas around the gaze point and scales their pixel radius by screen resolution~\citep{fov1,fov2,fov3,ilaslan2023gazevqa,peng2025eye}, 
we define the FOV region for each frame $I_t$ ($t \in [t_i^s, t_i^e]$) as the set of pixels within a fixed Euclidean distance from the fixation centroid:
\begin{equation}
\mathcal{R}_{i,t}^{\text{fov}}
= \left\{ (u,v) \in I_t \;\middle|\; \|(u,v) - (\bar{x}_i, \bar{y}_i)\|_2 \le \tau_{\text{fov}} \right\},
\end{equation}
where $(u,v)$ denotes pixel coordinates on frame $I_t$, $\tau_{\text{fov}}$ is the FOV radius, and $(\bar{x}_i, \bar{y}_i)$ is the fixation centroid defined in \Cref{eq:fixation}.
Following the normalized dispersion strategy commonly used for fixation modeling~\cite{ilaslan2023gazevqa}, 
we target a consistent FOV size across video sources, ensuring that FOV extraction captures comparable spatial extent regardless of video resolution.
The remaining portion of the frame defines the out-of-FOV region: 
\begin{equation}
\mathcal{R}_{i,t}^{\text{out}} = I_t \setminus \mathcal{R}_{i,t}^{\text{fov}}.
\end{equation}
Across the entire fixation interval, these per-frame regions form two 
time-indexed sequences:
\begin{equation}
\mathcal{R}_i^{\text{fov}}
= \{ \mathcal{R}_{i,t}^{\text{fov}} \}_{t = t_i^s}^{t_i^e},
\qquad
\mathcal{R}_i^{\text{out}}
= \{ \mathcal{R}_{i,t}^{\text{out}} \}_{t = t_i^s}^{t_i^e}.
\end{equation}

\textbf{Region-specific visual prompting.}  
To extract objects from each region, we employ a MLLM (InternVL3.5-38B~\citep{wang2025internvl3}) with spatially guided visual prompts.
For FOV region $\mathcal{R}_{i,t}^{\text{fov}}$,  
we crop a circular patch centered at $(\bar{x}_i, \bar{y}_i)$ with radius $r_{\tau_{fov}}$, 
and overlay a small red dot at the fixation center to explicitly indicate the user's point of gaze.  
This patch is then provided to the model as the inside-FOV visual input.
For the out-of-FOV region $\mathcal{R}_{i,t}^{\text{out}}$,  
we take the original frame $I_t$ and mask the circular FOV area by replacing all pixels within radius $\text{fov}$
with a solid black disk (see \Cref{fig:main_pipeline}). 
This removes all gaze-relevant content while preserving the surrounding contextual background, ensuring that the model extracts only non-attended objects.
Given the region sequences  
$\mathcal{R}_i^{\text{fov}}$ and  
$\mathcal{R}_i^{\text{out}}$,  
the MLLM outputs two corresponding object sets:
\begin{equation}
\mathcal{O}_i^{\text{fov}} = \mathrm{MLLM}(\mathcal{R}_i^{\text{fov}}), \qquad
\mathcal{O}_i^{\text{out}} = \mathrm{MLLM}(\mathcal{R}_i^{\text{out}}),
\end{equation}
where each detected object is accompanied by a brief caption used for downstream QA generation.

\textbf{Scanpath generation.}
Based on the fixation-level object sets,  
we construct a scanpath $\mathcal{S}$ that represents the temporal evolution of gaze-guided object observations.  
This scanpath preserves the temporal order of fixations, thereby representing how attention shifts across different spatial regions and semantic contexts in the video. 
Given a sequence of $N$ fixations $\mathcal{F}$ defined in \Cref{eq:fixation}, we define 
$\mathcal{S} = \{(\mathcal{O}_i^{\text{fov}}, \mathcal{O}_i^{\text{out}})\}_{i = 1}^N$ 
where each element corresponds to the FOV and out-of-FOV object sets extracted from the fixation clip $\mathcal{V}_i$.
This scanpath preserves the temporal order of fixations, thereby capturing how attention shifts across different spatial regions and semantic contexts in the video.  
For each fixation $f_i$, the corresponding object sets $\mathcal{O}_i^{\text{fov}}$ and $\mathcal{O}_i^{\text{out}}$ capture the local visual context. 
Across time, the sequence of fixations forms a gaze-conditioned trajectory.

\subsection{Human Verification} 
\label{sec:approach-human}
We finally employ human annotators to verify the generated scanpaths and extracted objects.
Annotators review each fixation episode and determine whether each $\mathcal{O}_i^{\text{fov}}$ and $\mathcal{O}_i^{\text{out}}$ should be included or excluded.
In particular, for $\mathcal{O}_i^{\text{fov}}$, annotators correct any mislabeled or missing objects and captions.
Only human-verified scanpaths and object sets are used for the final benchmark.
This verification process ensures high-quality yet scalable annotations, achieving an average correctness rate of approximately 83\%. 
Further details on human verification and annotation quality analysis
are provided in \Cref{sec:appendix:human-verification,sec:appendix:annotation-quality}, respectively.

\section{\ours{}}\label{sec:benchmark}
We introduce the \ours{} QA generation process and its task taxonomy spanning past, present, and proactive tasks, designed to quantitatively evaluate gaze-guided streaming video understanding.

\subsection{Benchmark Overview}

\textbf{Benchmark overview.}
\ours{} comprises 8,521 QA pairs from 285 videos.
As shown in \Cref{fig:tasks}, \ours{} encompasses 10 tasks spanning \colorbox{pastColor}{past}, \colorbox{presentColor}{present}, and \colorbox{proactiveColor}{proactive} understanding.
By incorporating gaze signals aligned with egocentric videos, \ours{} reflects realistic scenarios of human attention and interaction.
Detailed benchmark statistics are provided in \Cref{sec:appendix-data-detail}.

\textbf{Problem setup.} 
We formulate a gaze-guided streaming video understanding task, where the model must answer time-sensitive questions while observing temporally incoming frames and user gaze.  
At a query time $t_q$,  
the MLLM $\mathcal{M}$ receives the video–gaze context $(\mathcal{V}, \mathcal{G})$ (see \Cref{sec:approach-preproc}) within a specified temporal range and produces an answer $A$ for a question $Q$.  
We denote the short temporal window by $\omega$, where $\omega$ is 60 seconds following \citep{lin2024streamingbench,niu2025ovo}. 
Based on the accessible portion of the stream, we define three novel gaze-based streaming tasks as follows:

\vspace{-0.1cm}
\begin{equation*}
\begin{aligned}
\textbf{Past:} \quad 
& A = \mathcal{M}\!\left(Q;\, 
\mathcal{V}_{[0,\, t_q]},\,
\mathcal{G}_{[0,\, t_q]}\right), \\
\textbf{Present:} \quad 
& A = \mathcal{M}\!\left(Q;\, 
\mathcal{V}_{(t_q - \omega,\, t_q]},\,
\mathcal{G}_{(t_q - \omega,\, t_q]}\right), \\
\textbf{Proactive:} \quad 
& A = \mathcal{M}\!\left(Q;\, 
\mathcal{V}_{(t_q,\, \infty]},\,
\mathcal{G}_{(t_q,\, \infty]}\right).
\end{aligned}
\end{equation*}

\subsection{Task Taxonomy}
Given a scanpath $\mathcal{S}$ from \Cref{sec:approach-object-extraction}, we construct QA pairs that require spatio-temporal reasoning over gaze transitions. 
Examples of these tasks are illustrated in \Cref{fig:tasks}, and we describe their construction in detail in \Cref{sec:appendix:qa-generation}.

\subsubsection{\colorbox{pastColor}{Past} Task} \label{sec:benchmark-tasks-past}
Past tasks focus on temporal reasoning over gaze and capture the dynamic characteristics of the user’s gaze, modeling how attention shifts across objects over time. 

\begin{itemize}[leftmargin=1.5em]
\item \textbf{Non-Fixated Object Identification (NFI).} 
Evaluates implicit visual awareness by identifying objects that were visible but never directly fixated. 
For each timestamp $t$, we sample one never-gazed object from $\mathcal{O}_t^{\text{out}}$ as the correct answer and three visible objects from $\mathcal{O}_t^{\text{fov}}$ as distractors. 

\item \textbf{Object Transition Prediction (OTP).} 
Assesses temporal continuity in gaze behavior by predicting the next object to be fixated. 
Given the current fixation on object group $\mathcal{O}_i^{\text{fov}}$ within the scanpath $\mathcal{S}$, the correct answer is the next newly attended object $\mathcal{O}_{i+1}^{\text{fov}}$, while distractors are sampled from the global object pool excluding the current group.

\item \textbf{Gaze Sequence Matching (GSM).} 
Measures how well models capture human-like scanpath patterns through sequential gaze transitions. 
We extract triplets of consecutive fixation groups $\mathcal{O}_{i-1}^{\text{fov}} \rightarrow \mathcal{O}_{i}^{\text{fov}} \rightarrow \mathcal{O}_{i+1}^{\text{fov}}$ as correct transitions, while negative options include shuffled orderings of the same groups or randomly sampled triplets that preserve transition structure.

\item \textbf{Scene Recall (SR).} 
Tests contextual memory by recalling background objects previously visible during a fixation. 
At each fixation $\mathcal{O}_i^{\text{fov}}$, we sample the correct answer from background objects visible at other fixations 
$\mathcal{O}_{\setminus i}^{\text{out}} = \bigcup_{j \ne i} \mathcal{O}_j^{\text{out}}$ but not at the current one, with three visible background objects as distractors.
\end{itemize}

\subsubsection{\colorbox{presentColor}{Present} Task} \label{sec:benchmark-tasks-present}
Present tasks capture the user’s perceptual state and intent. 
We control question difficulty using both $\mathcal{O}_i^{\text{fov}}$ and $\mathcal{O}_i^{\text{out}}$.

\begin{itemize}[leftmargin=1.5em]
\item \textbf{Object Identification (OI, Easy/Hard).} 
Evaluates recognition of the currently attended object within the fixation region $\mathcal{O}_i^{\text{fov}}$. 
The correct answer corresponds to the fixated object itself. 
For the \textit{easy} setting, distractors are randomly sampled from earlier object pools, 
while for the \textit{hard} setting, distractors are chosen from visually similar $\mathcal{O}_i^{\text{out}}$ appearing in the same frame.

\item \textbf{Object Attribute Recognition (OAR).} 
Assesses fine-grained perceptual understanding by predicting visual attributes (e.g., color, shape, or texture) 
of the currently fixated object within $\mathcal{O}_i^{\text{fov}}$. 
To avoid overlapping or ambiguous attribute options, we use Qwen3-VL-30B~\citep{yang2025qwen3} to generate distinct and contextually consistent distractors.

\item \textbf{Future Action Prediction (FAP).} 
Models intent inference by predicting the user’s next action 
from the recent gaze-conditioned context $\{{\mathcal{O}_{i-k}^{\text{fov}}, \ldots, \mathcal{O}_{i}^{\text{fov}}}\}$,
where $k<i$ denotes the number of past fixation steps considered. 
Ground-truth future actions are derived from fine-grained human action captions aligned with gaze sequences in the original video sources. 
Semantically related but incorrect action descriptions are generated by Qwen3-VL-30B to serve as distractors.

\end{itemize}

\subsubsection{\colorbox{proactiveColor}{Proactive} Task} \label{sec:benchmark-tasks-future}
Proactive tasks go beyond passive perception by requiring model-initiated interventions~\citep{niu2025ovo}. 
The model must determine when to act (e.g., issue an alert), rather than merely predict future actions. 
Unlike prior proactive settings~\citep{zhang2025proactive,lin2024streamingbench,niu2025ovo}, we leverage gaze as an anticipatory signal for upcoming attention and object events.

\begin{itemize}[leftmargin=1.5em]
\item \textbf{Gaze-Triggered Alert (GTA).} 
Triggers an alert when the user gazes at a specified object within $\mathcal{R}_i^{\text{fov}}$. 
This task evaluates the model’s ability to indicate user's fixation.

\item \textbf{Object Appearance Alert (OAA).} 
Triggers an alert when the specified object appears in the peripheral region $\mathcal{R}_i^{\text{out}}$. 
This task assesses whether the model can proactively recognize and react to newly emerging objects in streaming video.
\end{itemize}

Overall, we filter out ambiguous questions through both Qwen3-VL-30B-based validation~\citep{yang2025qwen3} and human verification. 
Please refer to \Cref{sec:appendix:qa-generation} for more details. 

%% file: sec/4_exps.tex
\section{Experiments}

We conduct extensive experiments on \ours{} to evaluate
the capabilities of existing MLLMs in gaze-guided streaming
video understanding.
We first describe the experimental setup, then compare
model performance across past, present, and proactive tasks.
Finally, we provide additional analyses of gaze prompting,
reasoning strategies, and fixation extraction robustness.

\input{table/main_table}

\subsection{Experimental Setup}
\textbf{Baselines.}
We evaluate four categories of models on \ours{} in a zero-shot setting:
\textit{(i) closed-source MLLMs, (ii) gaze-based models,
(iii) open-source MLLMs}, and \textit{(iv) streaming MLLMs}.
Following prior streaming benchmarks~\citep{lin2024streamingbench,niu2025ovo},
we evaluate streaming models using native streaming inference,
while non-streaming models are evaluated via causal offline replay.
Both settings enforce strict causality, allowing models to access
only video and gaze observations available up to the current timestamp,
without access to future frames.
More details are provided in Appendix~\ref{sec:evaluation-detail}.

\textbf{Gaze input strategy.}  
To incorporate gaze into each model, we employ a visual prompting strategy inspired by~\citep{peng2025eye}.  
For each frame $I_t$, we overlay (i) a green dot indicating the instantaneous gaze coordinate $(x_t, y_t)$ and (ii) a red circular region representing FOV.  
We also prepend an instruction prompt that explicitly explains these cues to the model, e.g.,  
\textit{``The user's gaze center is indicated by a green dot. The red circle represents the area where the user's gaze is focused.''}  
Full prompts are provided in \Cref{sec:evaluation-detail}. 

\textbf{Evaluation metrics.}  
For past and present tasks, which are multiple-choice questions, we use Accuracy based on exact matching. 
For proactive tasks, we adopt a multi-triggering query protocol following~\citep{niu2025ovo,lin2024streamingbench} to simulate online decision-making.
At each timestamp, the model is prompted with the cumulative video observed so far and must answer \textit{“Yes”} if the target object has appeared in the field of view and \textit{“No”} otherwise. 
We use accuracy as the primary metric, as it penalizes premature or excessive triggers and thus best reflects the reliability of proactive alerts.
Additional recall results are provided in \Cref{sec:appendix-additional-analysis}.

\subsection{Overall Performance}

In \Cref{tab:main_table}, we report performances for all \ours{} tasks and baseline MLLMs.

\textbf{General-purpose MLLMs struggle to leverage gaze for long-term temporal reasoning.}
As shown in the Past columns (NFI, OTP, SR, and GSM) of
\Cref{tab:main_table}, general-purpose open-source MLLMs
generally outperform the gaze-based baseline AssistGaze,
yet remain substantially behind human performance.
For instance, Qwen2.5-VL achieves 0.518 on NFI and
0.450 on SR, compared to human scores of 0.700 and 0.903,
respectively.
These gaps suggest that current MLLMs struggle to
accumulate and utilize gaze-conditioned evidence across
extended temporal contexts.
Furthermore, their reliance on sparsely sampled frames
may limit their ability to capture continuous gaze transitions,
highlighting the challenges of long-term gaze-guided
temporal reasoning.

\textbf{Streaming MLLMs struggle with gaze-guided proactive tasks.}
As observed in prior works~\citep{fu2025vispeak,niu2025ovo}, dialogue-based streaming models such as VideoLLM-online~\citep{chen2024videollm} often fail to follow task instructions and instead produce generic descriptions (e.g., \textit{“You look at the camera.”}), due to their conversational training data.
In contrast, the non-dialogue streaming model, ViSpeak~\citep{fu2025vispeak} outperforms open-source non-streaming MLLMs on proactive tasks, demonstrating the benefit of frame-by-frame online processing.
This limitation arises from the absence of explicit mechanisms for linking gaze shifts to predictive temporal reasoning within current streaming architectures.

\subsection{Additional Analysis}

\input{table/prompt_ablation}

\textbf{Ablation of gaze input prompting and task-specific behavior.}
We analyze different strategies for incorporating gaze information into MLLMs using Qwen2.5-VL (7B), including textual prompts, visual prompts, and salience maps following~\citep{peng2025eye}. 
As shown in \Cref{tab:prompt_ablation}, the salience-map prompt achieves the best overall performance, particularly on present and proactive tasks, suggesting that spatially aggregated gaze cues are more effective than raw coordinates or frame-level overlays. However, none of the prompting strategies consistently outperform the gaze-free baseline, indicating that current MLLMs are not inherently equipped to interpret raw gaze signals. 

\begin{wraptable}{r}{0.48\textwidth}
\centering
\vspace{-10pt}
\tiny
\caption{\small
\textbf{Sensitivity analysis of fixation extraction thresholds.}
We report the average number of extracted fixations per video
(\# Fix.), the total number of generated QA pairs
(\# QA), and model ranking stability (Spearman's $\rho$)
across 13 MLLMs relative to the default configuration.
Light blue rows are the default threshold settings.
}
\label{tab:fixation_sensitivity}
\setlength{\tabcolsep}{3pt}
\renewcommand{\arraystretch}{0.7}
\resizebox{\linewidth}{!}{%
\begin{tabular}{l|c|ccc}
\toprule
Threshold &
Value &
\# Fix. &
\# QA &
$\rho$ \\
\midrule

& 0.10 & 10.3 & 257 & 0.972 \\
\rowcolor{lightblue} \multirow{-2}{*}{Radius ($r_{\mathrm{thresh}}$)} & 0.15 & 17.6 & 432 & 1.000 \\
& 0.30 & 32.0 & 442 & 0.996 \\

\midrule
& 1.0 s & 93.3 & 449 & 0.983 \\
\rowcolor{lightblue} \multirow{-2}{*}{Duration ($\tau_{\mathrm{dur}}$)} & 2.0 s & 17.6 & 432 & 1.000 \\
& 3.0 s & 5.1 & 205 & 0.933 \\

\midrule
\rowcolor{lightblue} Scene ($\tau_{\mathrm{scene}}$) & 0.90 & 17.6 & 432 & 1.000 \\
& 0.97 & 10.6 & 302 & 0.996 \\
& 0.99 & 4.7 & 119 & 0.900 \\

\bottomrule
\end{tabular}%
}
\vspace{-4.5pt}
\end{wraptable}

\paragraph{Sensitivity to Fixation Extraction Thresholds.}
We extract fixation intervals using the standard single-pass
I-DT growing-window algorithm, which jointly determines
fixation boundaries under spatial and minimum-duration constraints.
To assess the sensitivity of fixation extraction, we independently
vary the spatial radius ($r_{\mathrm{thresh}}$),
minimum fixation duration ($\tau_{\mathrm{dur}}$),
and scene-consistency ($\tau_{\mathrm{scene}}$)
thresholds while holding all other settings fixed.
For each configuration, we measure the average number of
extracted fixations per video, the resulting QA dataset size,
and model ranking stability across 13 evaluated MLLMs.
Ranking stability is quantified as
$\rho(\theta) = \mathrm{Spearman}(\theta,\theta_0)$,
where $\theta_0$ denotes the default configuration
and $\rho = 1.0$ indicates identical model rankings.
As shown in Table~\ref{tab:fixation_sensitivity},
threshold variations substantially affect the number of
extracted fixations and QA pairs, while model rankings remain
highly consistent ($\rho \geq 0.90$) across all settings.
These results demonstrate that benchmark comparisons are
robust to variations in fixation extraction thresholds.

\textbf{Gaze-based reasoning in MLLMs.}
To investigate how MLLMs utilize gaze signals during inference, we ablate the contributions of text-, gaze-, and visual-based reasoning using GPT-4o in \Cref{table:reasoning_ablation}. 
For textual reasoning, we apply standard chain-of-thought prompting (e.g., \textit{“Let’s think step by step”}).
For gaze-based reasoning, GPT-4o is first instructed to estimate the gaze point before generating an answer.
For visual reasoning, the model is guided to identify visible objects with bounding boxes and use this visual evidence when answering.
The best average performance is achieved when all reasoning strategies are combined.
However, their effectiveness varies across tasks: incorporating visual cues benefits Scene Recall (SR) by improving contextual grounding, but can hinder Non-Fixated Object Identification (NFI), where excessive focus on gaze regions may suppress exploration of unseen objects.
More details and qualitative examples are in \Cref{sec:appendix-additional-analysis}.

\input{table/reasoning}

\textbf{Proactive behavior of MLLMs.}
In \Cref{fig:proactive}, we analyze how offline MLLMs behave on \ours{} proactive remind tasks by measuring Type~1 (false positive) and Type~2 (false negative) error rates across each task. 
This evaluation reveals strong model-specific biases: InternVL3.5-8B consistently over-triggers with very high false positive rates, GPT4o deteriorates as tasks become harder with rising errors of both types, 
while Qwen2.5-VL-7B shows the most balanced pattern, remaining accurate in hard settings despite conservative behavior on easy ones. 
These results demonstrate that proactive gaze prediction is highly sensitive to model characteristics and underscore the need to examine error-type distributions when designing reliable gaze-based assistive systems.

%% file: table/main_table.tex
\begin{table*}[t]
  \centering
  \footnotesize
  \caption{\small\textbf{Comparison of various MLLMs across \ours{} tasks.}
  We evaluate four categories of models, reporting accuracy for past, present and proactive tasks. 
  Each section is sorted by overall performance, and the best per task across all models (excluding Human) is \textbf{bolded}.
  }
  \resizebox{\linewidth}{!}{
  \begin{tabular}{lcc|cccc|cccc|cc|c}
    \toprule
    \multirow{2}{*}{Method} &
    \multirow{2}{*}{Params} &
    \multirow{2}{*}{Frames} &
    \multicolumn{4}{c|}{\cellcolor[HTML]{E7F2DC}\textbf{Past}} &
    \multicolumn{4}{c|}{\cellcolor[HTML]{FBF5DC}\textbf{Present}} &
    \multicolumn{2}{c|}{\cellcolor[HTML]{FBE4E7}\textbf{Proactive}} &
    \multirow{2}{*}{\textbf{Overall}} \\

    & & &
    \cellcolor[HTML]{E7F2DC}NFI &
    \cellcolor[HTML]{E7F2DC}OTP &
    \cellcolor[HTML]{E7F2DC}SR &
    \cellcolor[HTML]{E7F2DC}GSM &
    \cellcolor[HTML]{FBF5DC}OI (E) &
    \cellcolor[HTML]{FBF5DC}OI (H) &
    \cellcolor[HTML]{FBF5DC}OAR &
    \cellcolor[HTML]{FBF5DC}FAP &
    \cellcolor[HTML]{FBE4E7}GTA &
    \cellcolor[HTML]{FBE4E7}OAA & \\

    \midrule
    Human & - & - &
    0.700 & 0.889 & 0.903 & 0.707 &
    0.960 & 0.920 & 0.800 & 0.840 &
    0.765 & 0.780 & 0.827 \\

    \midrule
    \rowcolor{lightgray}
    \multicolumn{14}{c}{\textit{Closed-Source MLLMs}}\\
    \midrule

    Claude Opus 4.8~\cite{anthropic2025claude4} & - & 16 &
    0.556 & 0.429 & 0.616 & 0.694 &
    0.720 & 0.713 & \textbf{0.611} & 0.424 &
    0.643 & 0.573 & \textbf{0.598} \\

    GPT-4o~\cite{openai2024gpt4technicalreport} & - & 16 &
    \textbf{0.601} & 0.449 & 0.535 & 0.580 &
    \textbf{0.729} & \textbf{0.730} & 0.596 & 0.370 &
    0.597 & 0.149 & 0.535 \\

    Claude Opus 4.6~\cite{anthropic2025claude4} & - & 16 &
    0.533 & 0.425 & 0.573 & 0.640 &
    0.479 & 0.525 & 0.343 & 0.413 &
    \textbf{0.646} & 0.662 & 0.524 \\

    Gemini 3.1 Pro~\cite{comanici2025gemini} & - & 16 &
    0.434 & 0.350 & 0.517 & 0.350 &
    0.640 & 0.581 & 0.443 & 0.395 &
    0.630 & \textbf{0.751} & 0.509 \\

    GPT-5.4~\cite{openai2025gpt5} & - & 16 &
    0.480 & 0.368 & \textbf{0.716} & 0.645 &
    0.492 & 0.486 & 0.360 & 0.249 &
    0.454 & 0.702 & 0.495 \\

    Gemini 2.5-pro~\cite{comanici2025gemini} & - & 16 &
    0.345 & 0.346 & 0.678 & \textbf{0.720} &
    0.466 & 0.423 & 0.363 & 0.159 &
    0.636 & 0.635 & 0.477 \\

    Claude Sonnet 4.6~\cite{anthropic2025claude4} & - & 16 &
    0.500 & \textbf{0.554} & 0.425 & 0.325 &
    0.521 & 0.533 & 0.561 & \textbf{0.439} &
    0.535 & 0.350 & 0.474 \\

    Claude Opus 4~\cite{anthropic2025claude4} & - & 16 &
    0.372 & 0.392 & 0.460 & 0.166 &
    0.431 & 0.436 & 0.430 & 0.351 &
    0.466 & 0.490 & 0.399 \\

    \midrule
    \rowcolor{lightgray}
    \multicolumn{14}{c}{\textit{GazeQA-based Fine-tuned Models}}\\
    \midrule

    AssistGaze~\cite{ilaslan2023gazevqa} & 26M & 32 &
    0.294 & 0.131 & 0.310 & 0.294 &
    0.278 & 0.250 & 0.109 & 0.254 &
    N/A & N/A & 0.223 \\

    \midrule
    \rowcolor{lightgray}
    \multicolumn{14}{c}{\textit{Open-Source MLLMs}}\\
    \midrule

    Qwen2.5-VL~\cite{wang2024qwen2vlenhancingvisionlanguagemodels} & 7B & Adaptive &
    0.518 & 0.350 & 0.450 & 0.483 &
    0.590 & 0.558 & 0.548 & 0.391 &
    0.486 & 0.407 & 0.478 \\

    Qwen3-VL~\cite{yang2025qwen3} & 8B & Adaptive &
    0.160 & 0.370 & 0.237 & 0.591 &
    0.603 & 0.649 & 0.534 & 0.427 &
    0.510 & 0.638 & 0.472 \\

    InternVL3.5~\cite{wang2025internvl3} & 8B & Adaptive &
    0.490 & 0.311 & 0.573 & 0.548 &
    0.627 & 0.628 & 0.466 & 0.372 &
    0.373 & 0.051 & 0.444 \\

    EgoGPT~\cite{yang2025egolife} & 7B & 32 &
    0.521 & 0.393 & 0.453 & 0.550 &
    0.690 & 0.620 & 0.304 & 0.368 &
    0.250 & 0.194 & 0.436 \\

    VITA 1.5~\cite{fu2025vita} & 7B & 16 &
    0.474 & 0.365 & 0.346 & 0.378 &
    0.455 & 0.396 & 0.437 & 0.370 &
    0.351 & 0.267 & 0.384 \\

    MiniCPM-V~\cite{yao2024minicpm} & 8B & 32 &
    0.430 & 0.374 & 0.354 & 0.296 &
    0.334 & 0.379 & 0.438 & 0.345 &
    0.480 & 0.216 & 0.365 \\

    Kangaroo~\cite{liu2024kangaroo} & 7B & 64 &
    0.363 & 0.365 & 0.319 & 0.275 &
    0.454 & 0.484 & 0.402 & 0.412 &
    0.242 & 0.198 & 0.351 \\

    \midrule
    \rowcolor{lightgray}
    \multicolumn{14}{c}{\textit{Open-Source Streaming MLLMs}}\\
    \midrule

    ViSpeak~\cite{fu2025vispeak} & 7B & 1 fps &
    0.463 & 0.358 & 0.417 & 0.473 &
    0.572 & 0.581 & 0.406 & 0.309 &
    0.635 & 0.458 & 0.467 \\

    TimeChat-Online~\cite{yao2025timechat} & 7B & 16 &
    0.500 & 0.206 & 0.441 & 0.274 &
    0.582 & 0.596 & 0.506 & 0.319 &
    0.376 & 0.197 & 0.400 \\

    Dispider~\cite{qian2025dispider} & 7B & 1 fps &
    0.366 & 0.365 & 0.381 & 0.263 &
    0.336 & 0.338 & 0.353 & 0.321 &
    0.252 & 0.261 & 0.323 \\

    Flash-VStream~\cite{zhang2024flashvstreammemorybasedrealtimeunderstanding} & 7B & 1 fps &
    0.249 & 0.202 & 0.336 & 0.220 &
    0.289 & 0.147 & 0.044 & 0.280 &
    0.443 & 0.217 & 0.243 \\

    VideoLLM-online~\cite{chen2024videollm} & 8B & 2 fps &
    0.000 & 0.000 & 0.000 & 0.000 &
    0.006 & 0.006 & 0.002 & 0.000 &
    0.458 & 0.333 & 0.080 \\

    \bottomrule
  \end{tabular}
  }
  \label{tab:main_table}
  \vspace{-0.15in}
\end{table*}

%% file: table/prompt_ablation.tex
\begin{wraptable}{r}{0.48\textwidth}
\centering
\vspace{-10pt}
\footnotesize
\caption{\small\textbf{Ablation of gaze input prompting.}
We report accuracy across different gaze input strategies.}
\label{tab:prompt_ablation}
\resizebox{\linewidth}{!}{%
\begin{tabular}{l|c c c c}
\toprule
 & 
\cellcolor[HTML]{E7F2DC}\textbf{Past} &
\cellcolor[HTML]{FBF5DC}\textbf{Present} &
\cellcolor[HTML]{FBE4E7}\textbf{Proactive} &
Avg. \\
\midrule
\rowcolor{lightgray}
Qwen2.5-VL (w/o gaze) 
& \textbf{0.423} & 0.500 & 0.384 & 0.446 \\
+ text prompt 
& 0.403 & 0.499 & 0.341 & 0.429 \\
+ visual prompt 
& 0.398 & 0.503 & 0.342 & 0.429 \\
+ salience map 
& 0.394 & \textbf{0.546} & \textbf{0.386} & \textbf{0.454} \\
\bottomrule
\end{tabular}%
}
\vspace{-5pt} 
\end{wraptable}

%% file: table/reasoning.tex
\begin{figure}[t]
\centering

\begin{minipage}[c]{0.49\textwidth}
\centering
\captionof{table}{\small\textbf{Ablation of gaze reasoning.}
We report accuracy when evaluating the effectiveness of text, gaze, and visual reasoning strategies using GPT-4o.}
\label{table:reasoning_ablation}
\setlength{\tabcolsep}{2.5pt} 
\renewcommand{\arraystretch}{0.9} 
\tiny
\resizebox{\columnwidth}{!}{%
\begin{tabular}{c c c | c c | c c | c}
\toprule
\multicolumn{3}{c|}{\textit{Reasoning}} &
\multicolumn{2}{c|}{\cellcolor[HTML]{E7F2DC}\textbf{Past}} &
\multicolumn{2}{c|}{\cellcolor[HTML]{FBF5DC}\textbf{Present}} &
\\
Text & Gaze & Visual & NFI & SR & OI & FAP & Avg. \\ 
\midrule
 &  &  & \textbf{0.52} & 0.60 & 0.64 & 0.41 & 0.542 \\
\ding{51} &  &  & 0.10 & 0.62 & 0.68 & \textbf{0.47} & 0.467 \\
\ding{51} & \ding{51} &  & 0.36 & 0.62 & 0.72 & 0.40 & 0.525 \\
\ding{51} & \ding{51} & \ding{51} &
0.44 & \textbf{0.66} &
\textbf{0.72} & 0.44 &
\textbf{0.565} \\
\bottomrule
\end{tabular}%
}
\end{minipage}
\hfill
\begin{minipage}[c]{0.47\textwidth}
\centering
\includegraphics[width=\columnwidth]{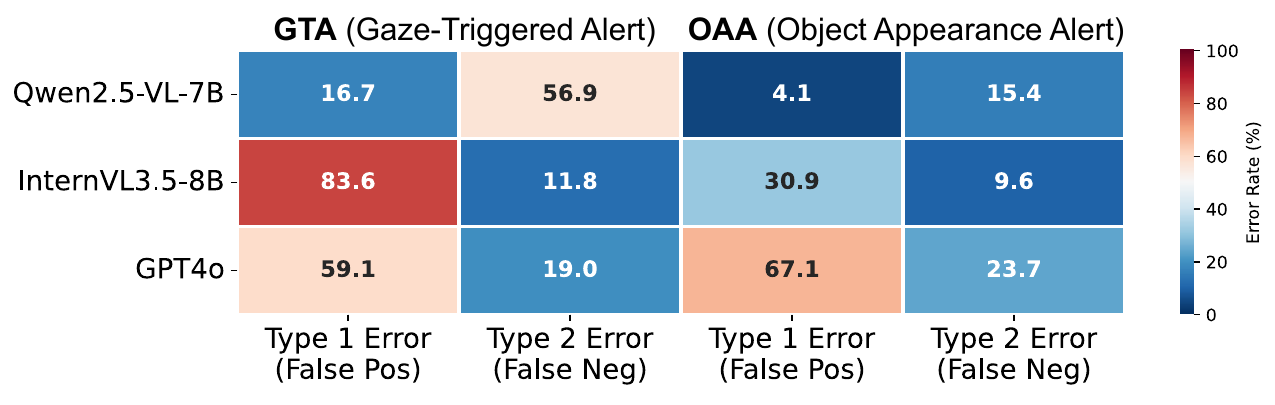}
\caption{\small\textbf{Offline MLLMs’ behavior on proactive tasks.} We evaluate Type 1 and Type 2 errors of Qwen2.5-VL, InternVL-3.5, and GPT4o.} 
\label{fig:proactive}
\end{minipage}
\vspace{-.15in}
\end{figure}

%% file: sec/5_conclusion.tex
\section{Conclusion}
In this work, we introduce \ours{}, the first benchmark for evaluating gaze-guided temporal and proactive reasoning in streaming video understanding. 
Our gaze–video QA pipeline provides human-aligned supervision across diverse past, present, and proactive tasks. 
Experiments reveal substantial gaps between state-of-the-art MLLMs and human annotators, especially in intent modeling and proactive prediction. 
We hope that \ours{} advances the development of more human-aligned, attention-aware streaming models.

\section{Limitations and Future Work}
Although \ours{} offers the first comprehensive benchmark for gaze-guided
streaming reasoning, some limitations suggest promising avenues for future
work. 
\ours{} spans ten heterogeneous tasks with inherently imbalanced sample sizes, and future work could explore curriculum or task-balanced training to improve uniformity across tasks. The benchmark is also grounded in manipulation-centric egocentric environments, leaving broader domains such as social interaction, navigation, and outdoor activities largely unexplored.

\section*{Acknowledgements and Disclosures}
This work was conducted during an internship at Adobe Research and was supported by Adobe Research. The authors declare no competing interests.

%% file: sec/supple.tex
\clearpage
\setcounter{page}{1}

\begin{center}
    {\Large\bfseries Supplementary Material for \ours{}\par}
    \vspace{1em}
\end{center}

\addcontentsline{toc}{section}{Appendix Table of Contents}
\startcontents[appendix]
\printcontents[appendix]{l}{1}{\setcounter{tocdepth}{2}}

\section{\ours{} Data Construction Details}\label{sec:appendix-data-construction}

In this section, we elaborate details about \ours{} data construction pipeline. 
We first introduce details of gaze projection (\cref{sec:appendix:gaze-proj}) during preprocessing~\cref{sec:approach-preproc} and fixation extraction (\cref{sec:appendix:fixation}). 
Then, we introduce object extraction details with prompt (\cref{sec:appendix:object-extraction}) and QA generation details by each task (\cref{sec:appendix:qa-generation}). 
Finally, we introduce human annotation process details (\cref{sec:appendix:human-verification}). 
We also provide the full data construction procedure in \cref{alg:pipeline}.

\subsection{Gaze Projection}\label{sec:appendix:gaze-proj}

To obtain frame-level gaze positions on the egocentric RGB video, we convert each 3D gaze ray into a 2D pixel coordinate using camera poses and calibrated intrinsics for the HoloAssist dataset~\citep{wang2023holoassist}. 
This method follows the hand–eye projection procedure implemented in official code.\footnote{\url{https://github.com/taeinkwon/PyHoloAssist/blob/main/hand_eye_project.py}}
Each gaze sample provides a world-coordinate origin $o(t)$ and direction $d(t)$, while the camera stream provides per-frame extrinsic matrices and intrinsics.

\input{table/algo_pipeline}

\textbf{Temporal alignment.}
All signals, including video frames, camera poses, and gaze samples, share a common timestamp domain.
For each video frame at time $t$, we select the nearest timestamped camera pose and gaze measurement:
\[
(o_t, d_t) = (o(t^*), d(t^*)) \quad\text{where}\quad
t^* = \arg\min_\tau |\tau - t|.
\]
This nearest-neighbor alignment ensures consistent pairing despite differing sampling rates.

\textbf{3D gaze point construction.}
Given the aligned 3D gaze ray, we form a virtual fixation point by moving a fixed distance $d_{\mathrm{eye}}$ along the normalized gaze direction:
\[
p_t^{\mathrm{world}} = o_t + d_{\mathrm{eye}}\, \frac{d_t}{\|d_t\|}.
\]
We treat $p_t^{\mathrm{world}}$ as the gaze target for frame $v_t$.

\textbf{World-to-camera transformation.}
Let $T_{\mathrm{cam}}^{\mathrm{world}}(t)$ be the camera pose associated with frame $v_t$.
We convert the world point into camera coordinates via
\[
p_{t,\mathrm{cam}} =
T_{\mathrm{axis}}
\left(T_{\mathrm{cam}}^{\mathrm{world}}(t)\right)^{-1}
\begin{bmatrix}
p_t^{\mathrm{world}} \\ 1
\end{bmatrix}.
\]

\textbf{Pinhole projection.}
With the intrinsic matrix
\[
K =
\begin{bmatrix}
f_x & 0   & c_x \\
0   & f_y & c_y \\
0   & 0   & 1
\end{bmatrix},
\]
we project the 3D camera-coordinate point $(X_t, Y_t, Z_t)$ onto the image plane:
\[
u_t = \frac{X_t}{Z_t}\, f_x + c_x, \quad
v_t = \frac{Y_t}{Z_t}\, f_y + c_y.
\]
A point is marked as in-frame if $(u_t, v_t)$ lies within the image boundaries.

\textbf{Output.}
For each frame $v_t$, the resulting projected coordinates $(u_t, v_t)$ define the 2D gaze position on the image plane.
We therefore set $(x_t, y_t) = (u_t, v_t)$ to construct the gaze trajectory 
$\mathcal{G} = \{(x_t, y_t)\}_{t=1}^{T}$ 
used in the main paper.
In addition, we store the original gaze validity flag and an in-frame indicator.
For visualization, we simply draw a dot at $(x_t, y_t)$ on the RGB frame.

\subsection{Fixation Extraction}\label{sec:appendix:fixation}
\paragraph{Handling short gaze interruptions.}
Human gaze often contains brief microsaccades or flicks that momentarily violate spatial stability. 
To avoid prematurely terminating a fixation, we permit short interruptions within a fixation interval.
For each timestamp $t$, if the gaze deviation exceeds the spatial threshold
($d_t > \tau_{\text{thresh}}$),
we examine the temporal gap between consecutive gaze samples:
\[
\Delta t = t - (t-1).
\]
If $\Delta t \le 200ms$ (an interruption threshold),
the deviation is treated as a transient flick rather than a fixation termination, 
and the fixation interval continues without interruption.

\textbf{Scene consistency.}
To ensure that each detected fixation corresponds to a visually coherent
segment, we further apply a scene-consistency filter based on
frame-wise Hue–Saturation histogram similarity. 
For a fixation interval $f_i = (\bar{x}_i, \bar{y}_i, t_i^s, t_i^e)$,
we uniformly sample $N$ frames from the corresponding video subsequence 
$\mathcal{V}_i = \{v_t\}_{t = t_i^s}^{t_i^e}$ 
and convert each sampled frame into the HSV color space. 
For every frame $v_t$, we compute a normalized 2D histogram over the
Hue–Saturation channels:
\[
H_t(h, s) = \mathrm{Hist}^{HSV}(v_t(h,s)), 
\quad h \in [0,180],\; s \in [0,256],
\]
with $\sum_{h,s} H_t(h,s) = 1$.

Next, we measure visual continuity between consecutive frames by computing
the histogram correlation using Pearson's coefficient:
\begingroup
\small
\begin{equation}
\label{eq:pearson-split}
\begin{aligned}
\rho(f_i(t), f_i(t+1))
= {} &
\frac{
\sum_{h,s}
(H_t(h,s)-\mu_t)\,(H_{t+1}(h,s)-\mu_{t+1})
}{
\sqrt{\sum_{h,s} (H_t(h,s)-\mu_t)^2}
}
\\[4pt]
& \times
\frac{1}{
\sqrt{\sum_{h,s} (H_{t+1}(h,s)-\mu_{t+1})^2}
}.
\end{aligned}
\end{equation}
\endgroup

where $\mu_t$ and $\mu_{t+1}$ denote the mean values of $H_t$ and
$H_{t+1}$. 
Collecting similarities across the entire episode gives
\begingroup
\small
\begin{equation}
\label{eq:s-collection}
S = 
\big\{
\rho(f_i(t_1), f_i(t_2)),\;
\dots,\;
\rho(f_i(t_{N-1}), f_i(t_N))
\big\}.
\end{equation}

\begin{equation}
\label{eq:smin}
S_{\min} = \min(S).
\end{equation}
\endgroup

A fixation is retained only if 
\[
S_{\min} \ge \tau_{\text{scene}},
\]
where $\tau_{\text{scene}}$ is an empirically set threshold (typically
$\tau_{\text{scene}} \approx 0.9$).  
Intervals failing to satisfy this constraint are considered to include
scene changes and are discarded.

\subsection{Object Extraction by Gaze Area.}\label{sec:appendix:object-extraction}

\textbf{FOV definition.}
When camera intrinsics are available~\citep{wang2023holoassist}, we compute the
horizontal field-of-view (HFOV) using the intrinsic matrix $K$ and frame image
width $W_{\text{pixels}}$:
\[
HFOV_{\text{rad}} = 
2 \arctan\!\left(\frac{W_{\text{pixels}}}{2 f_x}\right)
\]

\[
HFOV_{\deg} = HFOV_{\text{rad}} \cdot \frac{180}{\pi}
\]

If intrinsics are unavailable~\citep{egtea,grauman2022ego4d,huang2024egoexolearn},
we assume a canonical $HFOV_{\deg} = 90^\circ$.  
Following standard visual-angle conventions in \citep{fov1,fov2,fov3}, the
perifoveal region spans approximately $5^\circ$--$15^\circ$.  
For our FOV threshold, we adopt the upper perifoveal bound and set
$r_{\deg} = 15^\circ$, which specifies the angular radius that is later
converted into the pixel threshold $\tau_{\text{fov}}$.
We then convert this angular radius into a pixel radius via
\[
\text{px/deg} = \frac{W_{\text{pixels}}}{HFOV_{\deg}},
\qquad
\tau_{\text{fov}} = r_{\deg} \times \text{px/deg}.
\]
This $\tau_{\text{fov}}$ serves as the pixel threshold used in
\cref{sec:approach-object-extraction} to determine whether an object lies
inside or outside the FOV.

\textbf{Region-specific visual prompting.}  
To extract objects conditioned on human gaze, we design region-specific visual prompting strategies for InternVL-3.5 (38B).
As shown in \cref{fig:prompt-fov} and \cref{fig:prompt-out-fov}, we provide two distinct prompts: one for identifying objects within the user’s FOV and another for discovering objects located outside the FOV.

For the in-FOV setting, we provide the model with the fixation-aligned video clip and explicitly instruct it to (i) generate a brief scene-level description, (ii) identify the precisely gazed object indicated by the green dot, and (iii) describe additional objects located within the perifoveal region. The prompt enforces consistent naming by providing an object pool and requires detailed, attribute-rich captions for each recognized object.
For the out-of-FOV setting, we mask the central region of the frame and prompt the model to detect only objects outside the masked area.
The prompt includes similar constraints on naming, appearance grounding, and relational descriptions, while removing all references to the gazed object.
Both prompts further include human-centric rules—e.g., describing persons only when their full or upper body is visible, and requiring identity strings such as “person wearing [clothing description]” rather than generic labels.

These region-specific prompts guide the MLLM to reliably extract spatially grounded object sets that are aligned with gaze-conditioned context, enabling consistent FOV-aware and out-of-FOV object discovery across datasets.

\subsection{QA Generation and Filtering}\label{sec:appendix:qa-generation}

In this section, we provide details of QA generation and filtering for each \ours{} task. 
We also elaborate how we decide query time $t_q$ for each task. 

\subsubsection{Non-Fixated Object Identification (NFI)}
NFI evaluates whether a model can identify objects that were visible in the scene but never directly fixated by the user. 

\textbf{QA generation and filtering.}
We collect all FOV/out-FOV objects and define the scene-wide object pool as 
$\mathcal{A} = \bigcup_i (\mathcal{O}_i^{\mathrm{fov}} \cup \mathcal{O}_i^{\mathrm{out}})$.
Objects that were visible yet never fixated are defined as 
$\mathcal{N}_i = \mathcal{A} \setminus \mathcal{O}_i^{\mathrm{fov}}$.
For each index $i$, if $\mathcal{N}_i$ is non-empty, we sample the correct
answer from $\mathcal{N}_i$ and choose three distractors from the prefix FOV
object set $\mathcal{O}_{\le i}^{\mathrm{fov}}$.
Since NFI does not require visual verification, we rely directly on 
human-verified scanpaths and perform only formatting and consistency checks.

\textbf{Query time.} 
Let $f_i$ denote the set of distinct fixation objects observed up to timestamp
$i$ as defined in \cref{sec:approach-fixation-extraction}.  
We define the query timestamp $t_q$ as the earliest time at which three unique
fixation objects have been accumulated:
\[
t_q = \min \{\, i \mid |f_i| \ge 3 \,\}.
\]
At this point, we determine which objects have already been seen and which have not. 
To compute the response window, we gather all intervals in which the
distractor objects (i.e., previously fixated objects) appear within the FOV.
We then take the earliest and latest of these intervals and expand the window
by two seconds on both ends, producing a consistent response interval that
fully captures the temporal range in which distractor objects are visible.

\medskip

\subsubsection{Scene Recall (SR)}
SR evaluates contextual memory by determining which background object was previously visible but not visible during the current fixation. 

\textbf{QA generation and filtering.}
We construct the global background pool as 
$\mathcal{B} = \bigcup_i \mathcal{O}_i^{\mathrm{out}}$.
At fixation $i$, the correct answer is sampled from 
$\mathcal{B} \setminus \mathcal{O}_i^{\mathrm{out}}$, 
while distractors are chosen from the currently visible background objects 
$\mathcal{O}_i^{\mathrm{out}}$.
For filtering, we apply three consistency checks with Qwen3-VL: 
(1) the question must be clear and unambiguous,
(2) the selected answer must be an object that was never fixated, and 
(3) the answer object must appear in the video.

\textbf{Query time.}
We define the query timestamp $t_q$ as the moment when the user first begins to
fixate on any of the objects mentioned in the question. This ensures that the
model is evaluated exactly when the relevant scene context becomes available to
the user.

\subsubsection{Object Transition Prediction (OTP)}
OTP evaluates temporal continuity in gaze by predicting the next newly attended object. 

\textbf{QA generation and filtering.} 
For consecutive timestamps $(i, i+1)$, we use 
$\mathcal{O}_i^{\mathrm{fov}}$ 
and 
$\mathcal{O}_{i+1}^{\mathrm{fov}}$. 
The correct answer is the first object in $\mathcal{O}_{i+1}^{\mathrm{fov}}$ that does not belong to $\mathcal{O}_i^{\mathrm{fov}}$, corresponding to the newly attended object. 
Distractors are sampled from the global object pool, excluding both $\mathcal{O}_i^{\mathrm{fov}}$ and the correct answer. 
For filtering, since OTP transitions come directly from human-verified scanpaths, no additional visual filtering is required. 

\textbf{Query time.}
We define the query timestamp $t_q$ as the moment when the fixation on the
current object group $\mathcal{O}_i^{\mathrm{fov}}$ first begins, marking the onset of the transition.
To compute the response window, we use the full temporal span of the transition
event: we start two seconds before $\mathcal{O}_i^{\mathrm{fov}}$ becomes visible in the FOV and end
two seconds after the fixation on the newly attended object $\mathcal{O}_{i+1}^{\mathrm{fov}}$
concludes. 
This window provides continuous context across the transition,
capturing both the preceding fixation pattern and the complete fixation period
of the next object.

\medskip

\subsubsection{Gaze Sequence Matching (GSM)}
GSM evaluates whether models capture sequential gaze patterns across time. 

\textbf{QA generation and filtering.} 
We extract consecutive 3-step fixation sequences 
$[\mathcal{O}_i^{\mathrm{fov}}, \mathcal{O}_{i+1}^{\mathrm{fov}}, \mathcal{O}_{i+2}^{\mathrm{fov}}]$ 
and serialize them as 
$\mathcal{O}_i^{\mathrm{fov}} \rightarrow \mathcal{O}_{i+1}^{\mathrm{fov}} \rightarrow \mathcal{O}_{i+2}^{\mathrm{fov}}$
to form the correct sequence.
Three negative options are generated: one using a shuffled ordering of the same groups, and two using random groups sampled from the global object pool while preserving structural validity. 
To prevent ambiguous QA pairs, we automatically remove samples with Qwen3-VL~\citep{yang2025qwen3} where any distractor sequence overlaps with the correct sequence in more than 50\% of segments.

\textbf{Query time.}
We define the query timestamp $t_q$ as the end of the third fixation group
$\mathcal{O}_{i+2}^{\mathrm{fov}}$, which marks the completion of the three-step
transition sequence used for the question. 
The response window spans the entire
sequence interval, beginning from the onset of the first fixation group
$\mathcal{O}_i^{\mathrm{fov}}$ and extending to the end of the third group
$\mathcal{O}_{i+2}^{\mathrm{fov}}$. 
This window provides continuous temporal context for
evaluating whether the model can recognize the correct sequential gaze pattern
across all three segments.

\medskip

\subsubsection{Object Identification (OI, Easy/Hard)}
OI (Easy/Hard) evaluates whether models can correctly recognize the object currently fixated within the gaze-aligned cropped region $\mathcal{O}_i^{\mathrm{fov}}$.

\textbf{QA generation and filtering.} 
In the easy task, distractors are sampled from all objects appearing anywhere in the video, excluding the target object and all objects within the cropped FOV region.
In the hard task, distractors are drawn exclusively from $\mathcal{O}_i^{\mathrm{out}}$, i.e., objects visible in the same frame but outside the fixation region.
If fewer distractors than required are available, the corresponding sample is discarded to avoid trivial or ambiguous questions.

\textbf{Query time.}
Each prediction is evaluated within the response interval aligned with the
corresponding fixation episode. We therefore define the query timestamp
$t_q$ as the start time of the fixation.

\medskip

\subsubsection{Object Attribute Recognition (OAR)}
OAR evaluates fine-grained perceptual understanding by asking the model to infer a specific visual attribute (e.g., color, material, shape, texture, size, or state) of the object currently fixated. 

\textbf{QA generation and filtering.} 
To construct reliable attribute questions, we employ a constrained LLM-based template system.
We define a fixed dictionary of attribute types and pair each type with a corresponding question template (e.g., color → “What color is this object?”).
Given the caption of $\mathcal{O}_i^{\mathrm{fov}}$, Qwen3-VL-30B~\citep{yang2025qwen3} is prompted to:
(1) select the most appropriate attribute type,
(2) extract a concise correct answer grounded in the caption, and
(3) generate three visually plausible but incorrect distractors that do not overlap semantically with the correct attribute.
To remove ambiguous or semantically entangled distractors, we apply an additional verification step using Qwen3-VL~\citep{yang2025qwen3}, retaining only unambiguous multiple-choice sets.

\textbf{Query time.}
Similar to the OI task, we define the query timestamp
$t_q$ as the start time of the fixation.

\medskip

\subsubsection{Future Action Prediction (FAP)}
FAP evaluates proactive intent inference by predicting the upcoming
action based on the recent sequence of fixations.

\textbf{QA generation and filtering.}
For each fixation index $i$, we extract a 3-step fixation sequence capturing the short-term progression of the user’s attention. We then align this sequence with the official action annotations of the video and identify the earliest future action whose timestamp occurs at least a small temporal margin (3 seconds to 1 minute) after the fixation ends. This action is assigned as the correct label. Distractor options are sampled from other actions within the same video, while removing near-duplicate or semantically ambiguous actions to ensure clear discrimination.

\textbf{Query time.}
We define the query timestamp $t_q$ as the moment when the
fixation sequence ends, ensuring that the model makes predictions immediately
after observing all relevant gaze information. The response window begins two
seconds before the actual action timestamp, preventing temporal leakage while
anchoring the prediction to the correct action segment.

\begin{figure*}[t]
    \centering
    {
    \includegraphics[width=\textwidth]{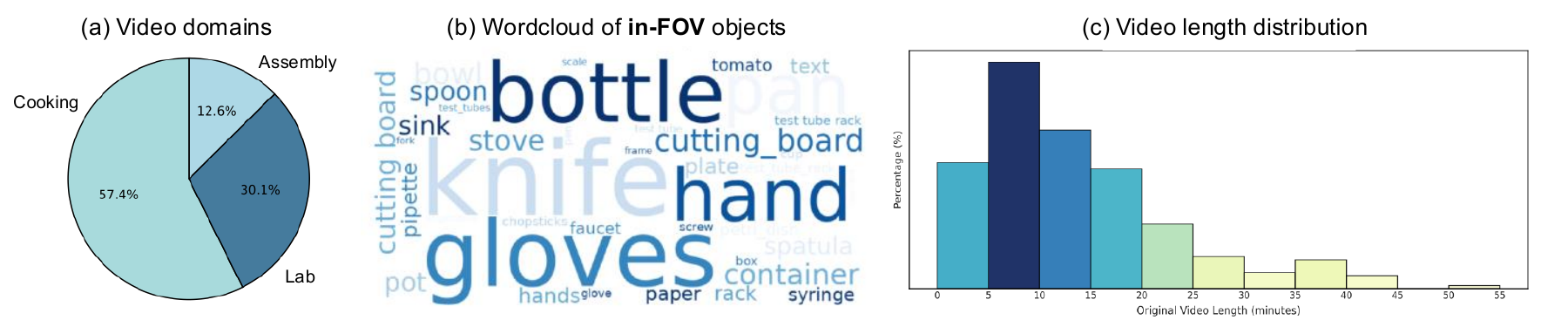}}
    \caption{
    \textbf{Data statistics of \ours{}.}
    We report domain proportion, video length, word cloud for FOV extracted objects. 
    }
    \label{fig:data_stat}
\end{figure*}

\begin{figure}[t]
\centering
\begin{minipage}[c]{0.48\textwidth}
\centering
\includegraphics[width=\columnwidth]{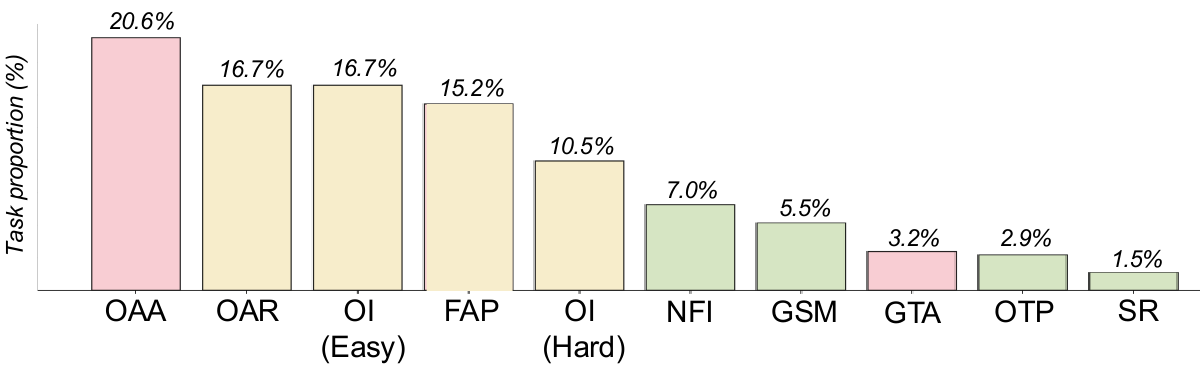}
\caption{\textbf{Task proportion.} We summarize the sample distribution for each task.}
\label{fig:task_proportion}
\end{minipage}
\hfill
\begin{minipage}[c]{0.48\textwidth}
\centering
\captionof{table}{\textbf{Human verification results per each video source.} We report object inclusion/modification ratio after human verification.}
\label{tab:human-verify-results}
\vspace{6pt}
\resizebox{\columnwidth}{!}{%
\begin{tabular}{l|cc}
\toprule
 & Inclusion Ratio (\%) & Modified Ratio (\%) \\ 
 \midrule
EGTEA-Gaze~\citep{egtea} & 81.77 & 6.89 \\
HoloAssist~\citep{wang2023holoassist} & 67.88 & 9.99 \\
EgoExoLearn~\citep{huang2024egoexolearn} & 84.31 & 7.24 \\ 
\bottomrule
\end{tabular}%
}
\end{minipage}
\end{figure}

\subsubsection{Gaze-Triggered Alert (GTA)}
GTA evaluates whether a model can proactively monitor the user’s fixation and
trigger an alert as soon as a specified target object enters the fixation region $\mathcal{R}_i^{\mathrm{fov}}$.
For each video, we first gather all objects that were directly fixated at least once, excluding background furniture or static structures using Qwen3-based filtering~\citep{yang2025qwen3}.
We then sample multiple evaluation timestamps relative to the object’s first fixation moment (e.g., $t-20$, $t-10$, $t$, $t+10$, $t+20$ seconds), following the multi-query design of OVO-Bench~\citep{niu2025ovo}.
At each timestamp, we determine whether the user is currently fixating on the object and assign a binary label accordingly (type~1 for fixation, type~0 for non-fixation).
For quality control, we manually verify the fixation labels to ensure accuracy.

\medskip

\subsubsection{Object Appearance Alert (OAA)}
OAA evaluates whether the model can proactively detect objects that appear in the peripheral region $\mathcal{R}_i^{\mathrm{out}}$, i.e., objects that are visible in the frame but have not yet entered the user’s fixation.
We extract all $\mathcal{O}_i^{\mathrm{out}}$ at each timestamp, removing background furniture or static structures using Qwen3-based filtering~\citep{yang2025qwen3}.
For each selected object, we generate evaluation timestamps (e.g., $t-20$, $t-10$, $t$, $t+10$, $t+20$ seconds) relative to its appearance, using offsets centered around the event boundary.
At each timestamp, we determine whether the object is currently visible in the frame but outside the fixation region, assigning type~1 when present and type~0 otherwise.
For quality control, we use Qwen3-VL~\citep{yang2025qwen3} to confirm that the object indeed appears within the 10-second evaluation window and additionally perform manual verification to ensure accuracy.

\subsection{Human Verification}\label{sec:appendix:human-verification}
\begin{figure*}[t]
    \centering
    {
    \includegraphics[width=\textwidth]{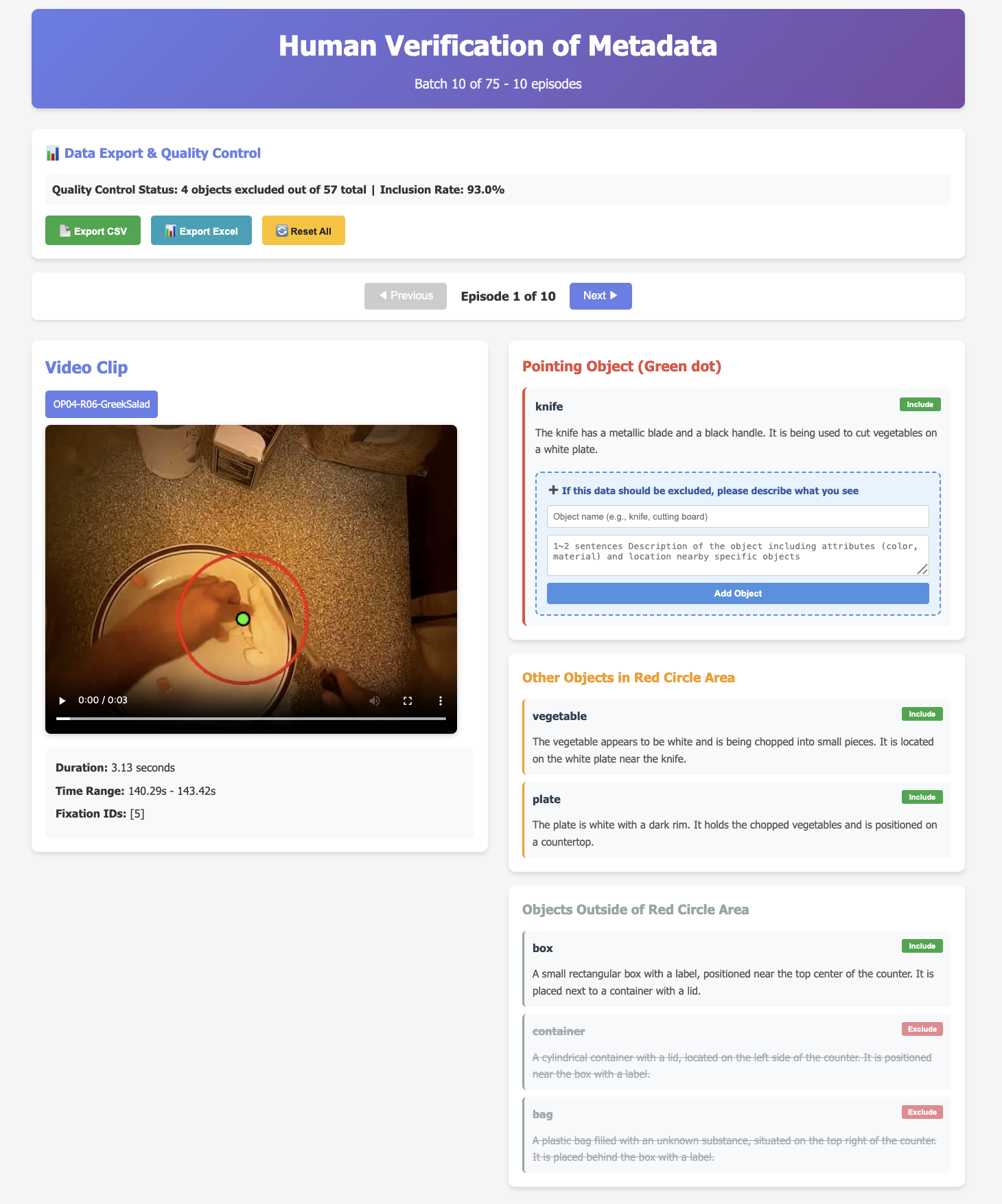}}
    \caption{
    \textbf{HTML for human verification of \ours{} data construction.}
    }
    \label{fig:human_anno_html}
\end{figure*}

\begin{figure*}[t]
    \centering
    {
    \includegraphics[width=\textwidth]{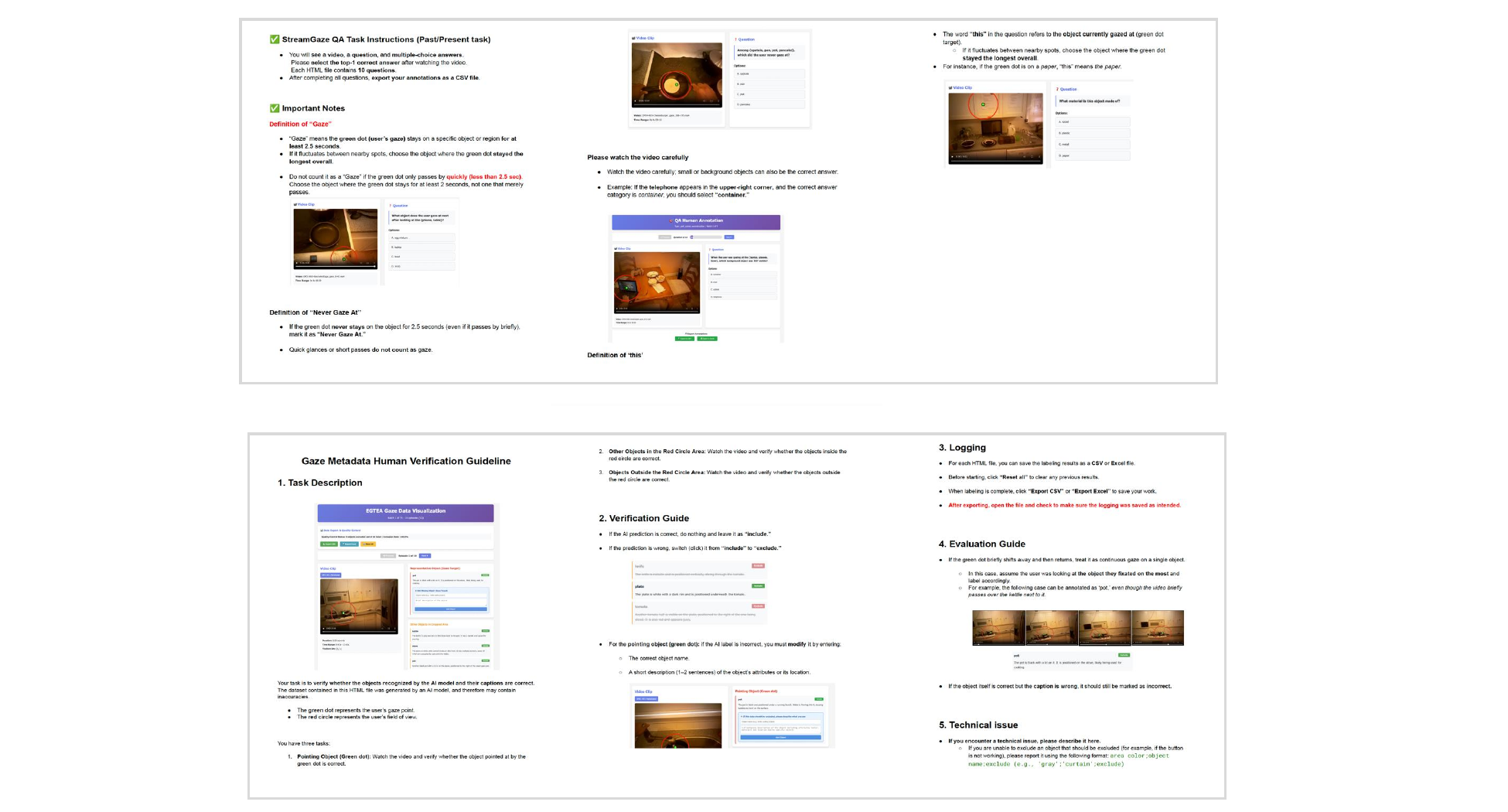}}
    \caption{
    \textbf{Instructions provided to human annotators.}
    }
    \label{fig:instruction}
\end{figure*}

As shown in \cref{fig:human_anno_html}, we employ three annotators and provide them with an HTML-based interface for verification. Each annotator is shown the fixation-level video clip along with the objects extracted by the MLLM: the precisely gazed object (marked by a green dot), other objects within the FOV region, and out-of-FOV objects. (Details of the provided instructions are shown in \cref{fig:instruction}.)
Annotators then determine whether each object should be included or excluded based on the visual evidence. For the gazed (pointing) object, annotators additionally correct or rewrite its identity and detailed caption when necessary.
We report the inclusion ratio and modification ratio in \cref{tab:human-verify-results}.
The inter-annotator agreement, measured using Fleiss’ Kappa, is 0.60, indicating consistency among annotators.

\subsection{Annotation Quality Analysis}
\label{sec:appendix:annotation-quality}

\paragraph{Object Annotation Verification.}
To ensure annotation reliability, we incorporate human verification
into the object annotation pipeline.
Specifically, object annotations proposed by InternVL3.5-38B
undergo manual verification prior to QA generation.
The object inclusion and independent re-annotation ratios
are reported in Table~\ref{tab:human-verify-results}.
As the majority of QA pairs are constructed directly from
these verified annotations, this verification step serves
as a key quality-control mechanism for benchmark construction.

\paragraph{QA-Level Quality Audit.}
Beyond object-level verification, we conduct an additional
manual audit to assess the quality of the generated QA pairs.
We examine 50 QA pairs per task (500 in total) along three
dimensions: (1) answer correctness, (2) distractor quality,
and (3) representative failure modes.
Answer correctness assesses the consistency of ground-truth
answers with the corresponding video and gaze observations,
while distractor quality measures whether incorrect options
are unambiguously distinguishable from the correct answer.
As summarized in Table~\ref{tab:qa_quality_audit},
the resulting QA pairs achieve \textbf{97.6\% answer correctness}
and \textbf{98.0\% distractor quality}.
The remaining errors primarily involve occasional annotation
inaccuracies and inherently ambiguous cases, including
uncertain object transitions, fixation ordering, and
temporal boundaries.
Overall, these findings provide empirical evidence for
the reliability of our annotation pipeline and the quality
of the resulting benchmark.

\begin{table}[t]
    \centering
    \scriptsize
    \renewcommand{\arraystretch}{0.90}
    \caption{
    \textbf{QA-level annotation quality audit.}
    We manually evaluate 50 QA pairs per task for
    answer correctness and distractor quality.
    Representative failure modes are reported for each task.
    N/A indicates tasks without distractor options.
    }
    \vspace{.3cm}
    \label{tab:qa_quality_audit}
    \resizebox{\columnwidth}{!}{
    \begin{tabular}{lcc l}
        \toprule
        \textbf{Task} &
        \textbf{Answer Correctness (\%)} &
        \textbf{Distractor Quality (\%)} &
        \textbf{Representative Failure Mode} \\
        \midrule

        \cellcolor[HTML]{E7F2DC} NFI & 100 & 98 &
        Incorrect never-fixated label \\

        \cellcolor[HTML]{E7F2DC} OTP & 96 & 100 &
        Ambiguous object transition \\

        \cellcolor[HTML]{E7F2DC} SR & 100 & 98 &
        Background object misannotation \\

        \cellcolor[HTML]{E7F2DC} GSM & 96 & 100 &
        Ambiguous fixation ordering \\

        \cellcolor[HTML]{FBF5DC} OI & 100 & 98 &
        Confusable object distractor \\

        \cellcolor[HTML]{FBF5DC} OAR & 94 & 96 &
        Ambiguous object state \\

        \cellcolor[HTML]{FBF5DC} FAP & 98 & 96 &
        Ambiguous action onset \\

        \cellcolor[HTML]{FBE4E7} GTA & 98 & N/A &
        Multiple valid object instances \\

        \cellcolor[HTML]{FBE4E7} OAA & 96 & N/A &
        Ambiguous appearance timing \\

        \midrule
        \textbf{Overall} &
        \textbf{97.6} &
        \textbf{98.0} &
        -- \\
        \bottomrule
    \end{tabular}
    }
\end{table}

\section{\ours{} Evaluation Details}\label{sec:evaluation-detail}

\paragraph{Baselines.}
We evaluate four categories of models on \ours{} in a zero-shot setting:
(1) \textit{Closed-source MLLMs}, including
GPT-4o~\citep{openai2024gpt4technicalreport},
GPT-5.4~\citep{openai2025gpt5},
Claude Sonnet 4.6, Claude Opus 4, 4.6, and 4.8~\citep{anthropic2025claude4},
and Gemini 2.5 Pro~\citep{comanici2025gemini} and Gemini 3.1 Pro~\citep{comanici2025gemini};
(2) \textit{Gaze-based models}, including
AssistGaze~\citep{ilaslan2023gazevqa};
(3) \textit{Open-source MLLMs}, including
Qwen2.5-VL~\citep{bai2025qwen25}, Qwen3-VL~\citep{yang2025qwen3},
InternVL3.5~\citep{wang2025internvl3},
MiniCPM-V~\citep{yao2024minicpm},
Kangaroo~\citep{liu2024kangaroo},
EgoGPT~\citep{yang2025egolife},
and VITA-1.5~\citep{fu2025vita};
and (4) \textit{Streaming MLLMs}, including
Flash-VStream~\citep{zhang2024flashvstreammemorybasedrealtimeunderstanding},
VideoLLM-Online~\citep{chen2024videollm},
Dispider~\citep{qian2025dispider},
ViSpeak~\citep{fu2025vispeak},
and TimeChat-Online~\citep{yao2025timechat}.

Following prior streaming benchmarks~\citep{lin2024streamingbench,niu2025ovo},
we evaluate streaming models via native streaming inference,
processing frames incrementally as they arrive,
while non-streaming models are evaluated via causal offline replay.
Both evaluation modes follow the same strictly causal protocol:
at each timestamp, models can access only video and gaze observations
available up to that point, without access to future frames.

\paragraph{Multiple-choice question evaluation.}
For all past and present tasks, we construct the model's input clip according to the prior streaming benchmarks~\citep{niu2025ovo,lin2024streamingbench}. 
Past tasks receive the full history from the beginning of the video up to the query timestamp $[0, t_q]$, enabling long-range reasoning. 
Present tasks use a 60-second sliding window $[t_q-60, t_q]$ (clamped to start at 0), reflecting real-time processing with limited recent context. 
All these tasks are evaluated as four-way multiple-choice questions. 
Because models may output answers in diverse natural-language formats, we extract the predicted option by deterministic parsing rules, prioritizing concise outputs such as a final standalone letter or expressions like ``the answer is B,'' while also supporting formats such as ``C. pot.''
Accuracy is reported as the proportion of correctly answered questions over the total number of questions for each task.

\input{table/sft}

\paragraph{Proactive evaluation.}
Following OVO-Bench~\citep{niu2025ovo}, we adopt a \textit{fair and unified evaluation} protocol that supports both \textit{offline} VideoLLMs (\eg, Qwen2.5-VL) and \textit{online} models. We formulate proactive tasks as a time-indexed binary decision problem, evaluated at multiple temporal checkpoints around a target event. Specifically, we define a set of checkpoints $r_t$ centered on the first fixation time of the target object (e.g., $t-20$, $t-10$, $t$, $t+10$, $t+20$ seconds).
At each checkpoint $r_t$, the model receives the cumulative video prefix $[0, r_t]$ and outputs a binary \textit{yes/no} decision indicating whether the target object lies within the fixation region. Each checkpoint is labeled as positive if the object is fixated and negative otherwise, and overall accuracy is computed across all checkpoints. This formulation evaluates whether a model can continuously track gaze signals and trigger alerts with precise temporal alignment.

Following prior streaming benchmarks (\eg, OVO-Bench), we define \textit{proactive} tasks as model-initiated interventions that require deciding \emph{when} to act based on evolving attention or environmental signals, rather than merely predicting future events. In this context, GTA and OAA extend beyond passive perception: both require continuous monitoring of gaze or scene dynamics and timely \textit{alert} generation without an explicit user query.

\section{Additional Analysis}\label{sec:appendix-additional-analysis}
\subsection{Fine-tuning Results}
To better understand the underlying failure of MLLMs and assess whether performance can be improved through gaze-supervised learning, we conduct LoRA-based fine-tuning experiments. 
Specifically, we fine-tune ViSpeak~\citep{fu2025vispeak} on our gaze-guided training set to quantify the potential performance gains achievable through targeted adaptation.

\textbf{Fine-tuning dataset.}
We construct four variants of training data for our fine-tuning experiments, drawing from both external resources and our own automatically generated data. 
First, we include external gaze-based QA datasets such as HD-EPIC~\citep{hdepic} (2k) and EgoGazeVQA~\citep{peng2025eye} (1.2k), which provide high-quality gaze–question–answer supervision despite not being strictly streaming-based. 
Second, we leverage QA pairs generated by our data construction pipeline. Because the scanpath inclusion ratio between our pipeline and human annotators is approximately 78\% (see \cref{tab:human-verify-results}), we further apply the same pipeline to EgoExoLearn~\citep{huang2024egoexolearn} and Ego4D-Gaze~\citep{grauman2022ego4d}, which do not overlap with \ours{} video sources, to automatically synthesize additional \ours{}-style training data. This serves as a form of data augmentation, expanding coverage and increasing the diversity of gaze-conditioned supervision.

\textbf{Implementation details.}  
The base model is ViSpeak, built on Qwen2.5-Instruct with an InternViT-300M vision tower and a frozen VITA-1.5 audio encoder. We enable LoRA with rank~128 and $\alpha=256$, and train the MLP-based multimodal projector with a learning rate of $1\!\times\!10^{-5}$. We use a batch size of 2 per device (gradient accumulation 2), a maximum sequence length of 5500 tokens, and train for 2 epochs.

\begin{figure}[h]
\centering
\begin{tcolorbox}[
    colback=gray!10,
    colframe=black!80,
    title={Negative sampling for proactive task fine-tuning
    },
    width=\columnwidth,
    boxrule=0.8pt
]
\scriptsize
\begin{verbatim}
{
"conversations": [
{"from": "human", 
"value": "Monitor and alert when I gaze <screen>", 
"time": 91.0},
{"from": "gpt", "value": "", "time": 2783.0},  
{"from": "gpt", "value": "", "time": 2793.0},  
{"from": "gpt", 
"value": "You are now gazing <screen>.", 
"time": 2803.0}  
],
  "proactive": true
}
\end{verbatim}
\end{tcolorbox}
\caption{\textbf{Example of negative sampling strategy for proactive task fine-tuning.} We apply negative sampling to fine-tune ViSpeak.}
\label{fig:negative_sampling}
\end{figure}

\textbf{Proactive task fine-tuning.}
For proactive reminder tasks, we fine-tune the model using a negative–positive sampling strategy together with ViSpeak's informative head. As illustrated in \cref{fig:negative_sampling}, all timestamps before the target object's first fixation are treated as \emph{negative} instances, where the model is trained to output an empty response so that it learns when it should remain silent. At the first fixation timestamp, a \emph{positive} instance is provided and the model is trained to generate the alert message, teaching it the exact moment a notification should be triggered. Each training example is formatted as a short multi-turn sequence in which early turns contain empty responses and the first positive turn contains the alert (e.g., ``You are now gazing <screen>."), clearly marking the transition from non-trigger to trigger conditions.

\textbf{Quantitative results.} 
As shown in \cref{tab:sft_result}, training only on external gaze QA datasets lowers the overall accuracy, indicating that these datasets alone do not transfer effectively to the streaming gaze reasoning setting. In contrast, adding automatically generated \ours{}-style data leads to clear improvements. EgoExoLearn is particularly helpful because it is in-domain with respect to egocentric manipulation scenes, making its synthesized gaze supervision more aligned with the target distribution. Incorporating Ego4D-Gaze provides additional gains, and combining all data sources yields the best overall accuracy of 0.521. However, because \ours{} covers a wide range of tasks, the resulting data distribution is inherently imbalanced, so performance does not increase uniformly across all tasks. Overall, these results show that in-domain synthetic gaze supervision is essential for improving ViSpeak, while external datasets alone are insufficient for this setting.

\begin{figure*}[t]
    \centering
    {
    \includegraphics[width=\textwidth]{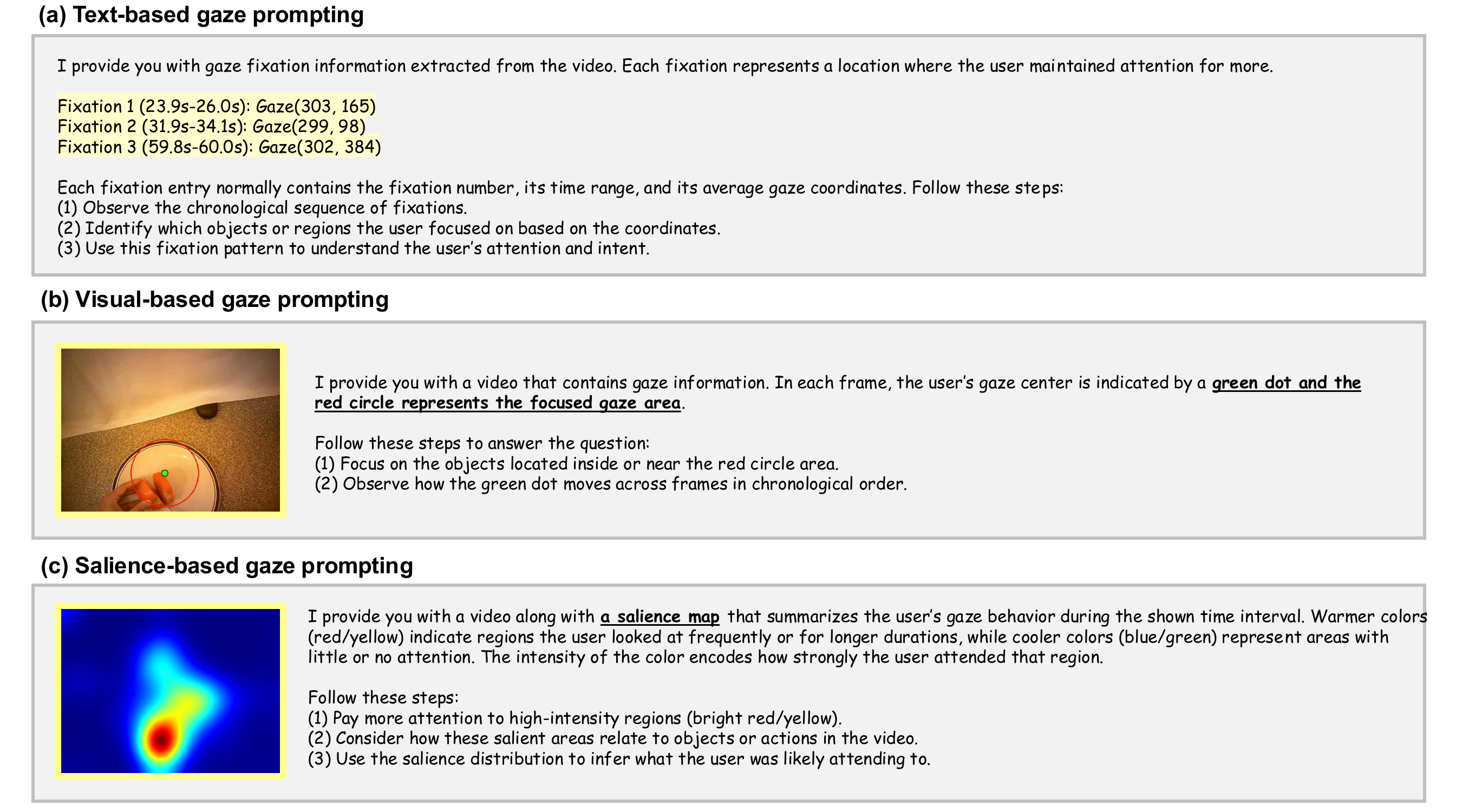}}
    \caption{
    \textbf{Gaze prompting strategies for \ours{}}
     We adopt three strategies for inputting gaze information to MLLMs.
    }
    \label{fig:prompting_method}
\end{figure*}

\begin{figure}[t]
    \centering
    {
    \includegraphics[width=\columnwidth]{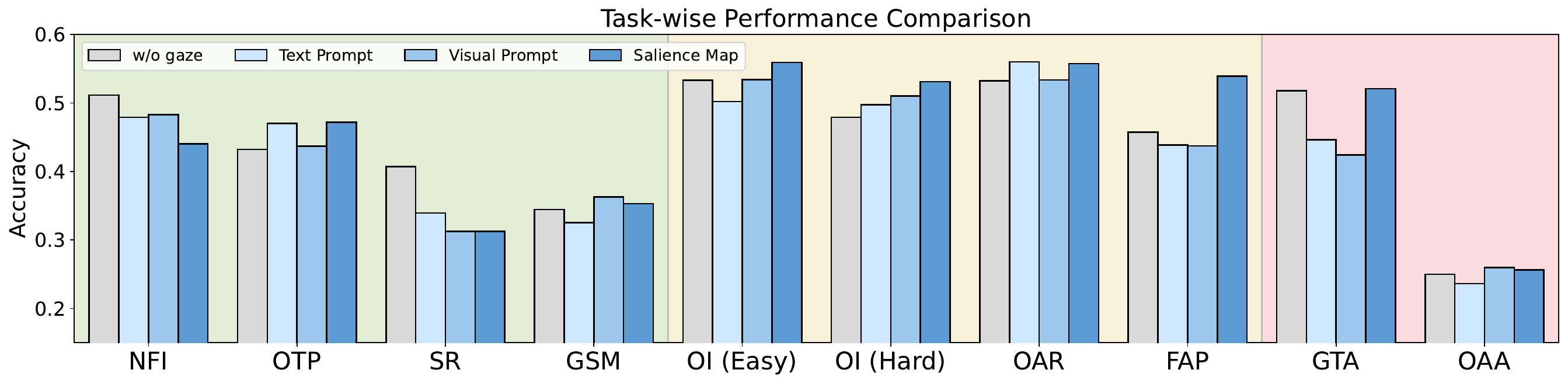}}
    \caption{
    \textbf{Task-wise effect of gaze input strategies on \ours{}.}
    Different tasks respond unevenly to textual, visual, and salience-map prompts, revealing strong task-specific behavior in streaming gaze understanding.
    }
    \label{fig:task_diff}
\end{figure}

\input{table/prompt_ablation_total}

\subsection{Details of Gaze Input Prompting}

We provide each gaze-prompting instruction in \cref{fig:prompting_method}, inspired by \citep{peng2025eye}, to obtain the Qwen2.5-VL performances reported in \cref{tab:prompt_ablation}. 
For the text gaze prompting setting, we extract the user’s fixation center prior to the query time and embed the corresponding fixation coordinates into the textual prompt. To construct a salience-map prompt, we follow Algorithm~1 from Appendix~B of \citep{peng2025eye} without modification, compressing the entire gaze trajectory into a single salience heatmap. We then prepend this salience map to the input video frames so that the MLLM jointly attends to both the visual stream and the gaze-derived salience distribution.
\cref{tab:prompt_ablation_total} also provide detailed performance per each task on GPT-4o, InternVL-3.5 and ViSpeak.

\subsection{Details of Gaze-based Reasoning}

\begin{figure*}[h]
    \centering
    {
    \includegraphics[width=\textwidth]{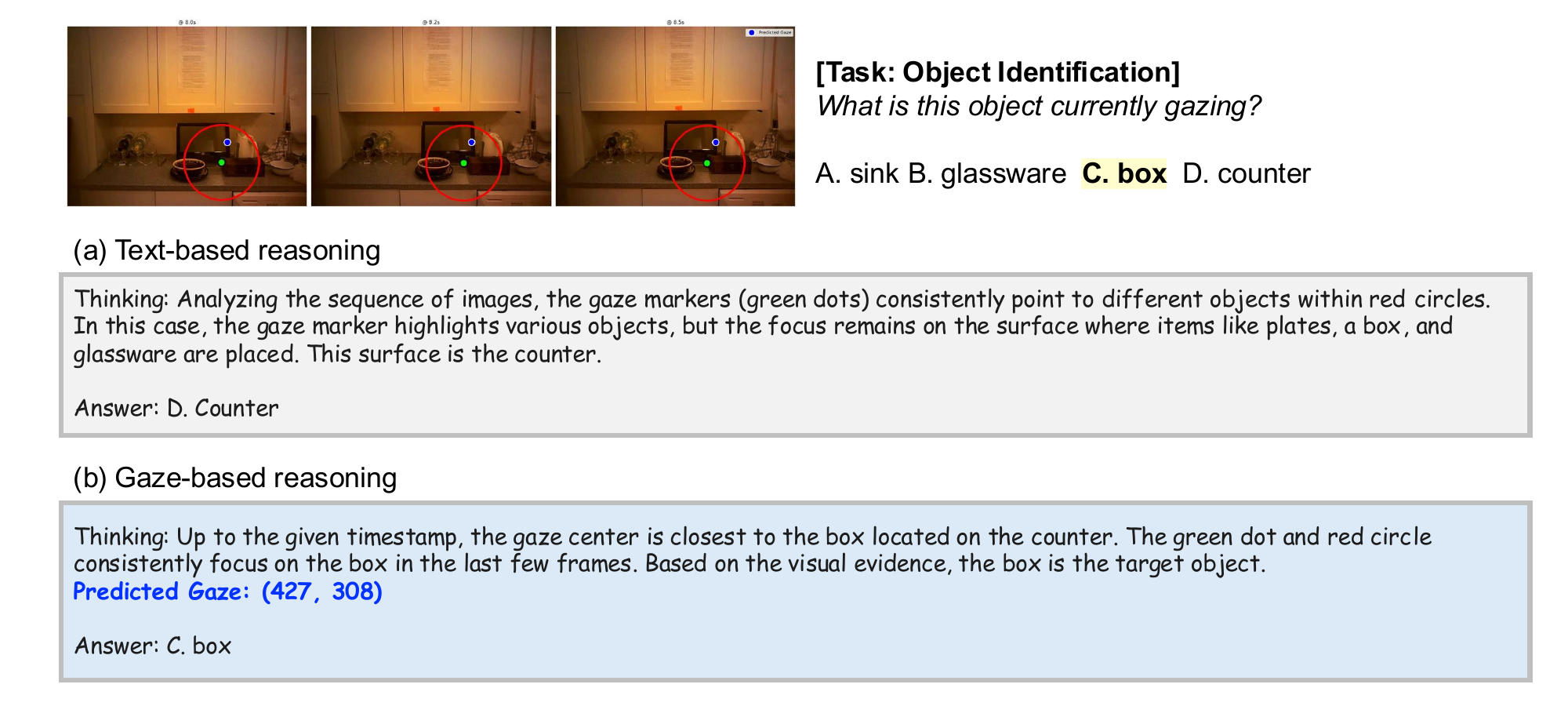}}
    \caption{
    \textbf{Qualitative comparison with text reasoning and gaze-based reasoning.}
    We visualize examples from the scene reconstruction task comparing (a) text and (b) gaze-based reasoning. 
    }
    \label{fig:cot1}
\end{figure*}

\begin{figure*}[h]
    \centering
    {
    \includegraphics[width=\textwidth]{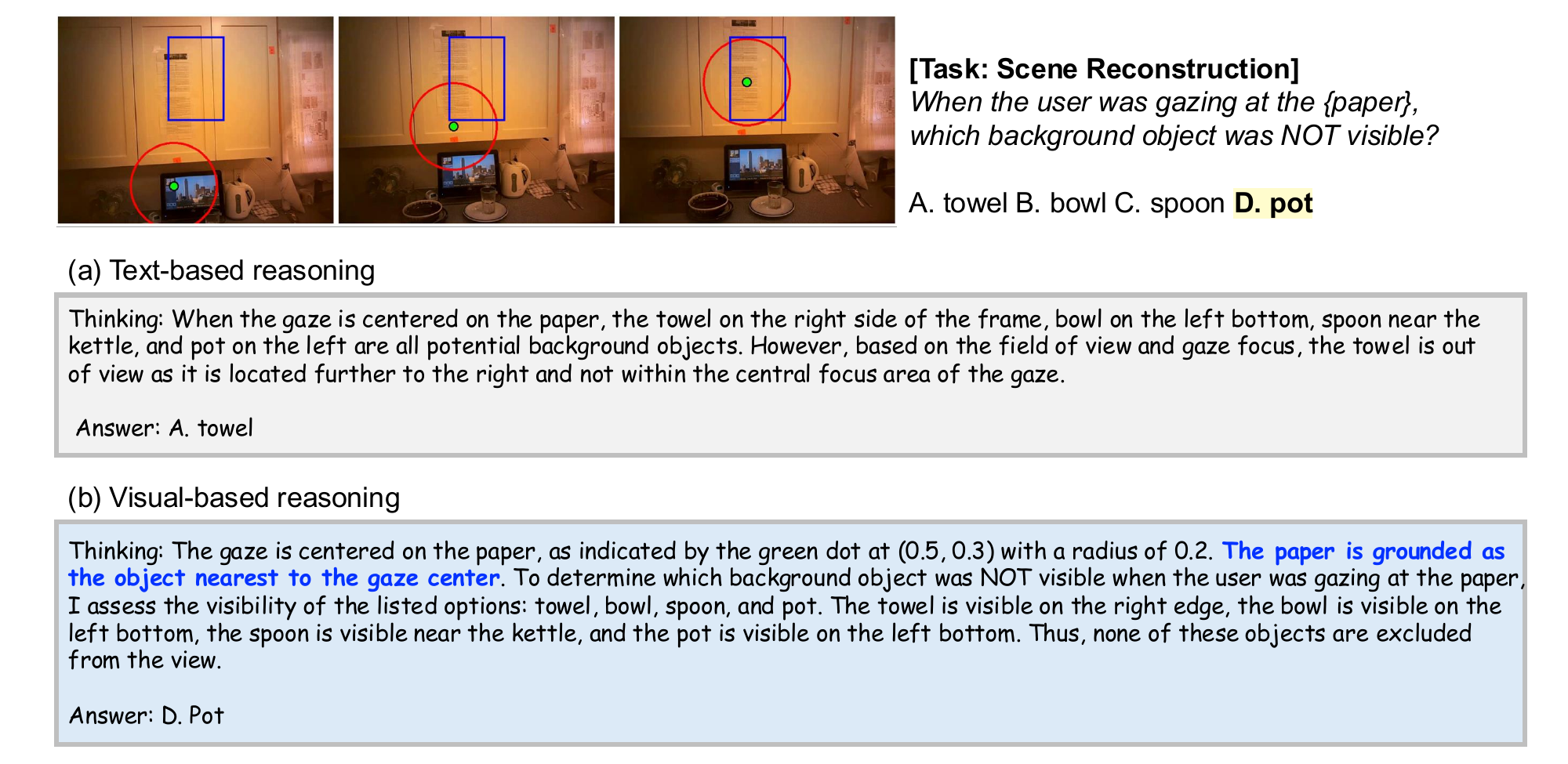}}
    \caption{
    \textbf{Qualitative comparison with text reasoning and visual-based reasoning.}
    We visualize examples from the object identification task comparing (a) text and (b) visual-based reasoning. 
    }
    \label{fig:cot2}
\end{figure*}

We illustrate the input formats for gaze-, text-, and visual-based reasoning in \cref{fig:gaze-reasoning-input,fig:text-reasoning-input,fig:visual-reasoning-input}.
We further provide qualitative comparisons between text-only reasoning and gaze- or visual-guided reasoning in \cref{fig:cot1,fig:cot2}.
In \cref{fig:cot1}, the gaze-based reasoning correctly identifies the attended object by leveraging the fixation location, whereas the text-only reasoning fails due to relying solely on language priors without visual grounding.
In \cref{fig:cot2}, the visual-based reasoning succeeds by using the detected object region to determine which background item was outside the user’s field of view, while the text-only reasoning struggles because it lacks explicit spatial grounding.

\subsection{Effect of Fixation Filtering}\label{sec:appendix:fixation-filtering}

\begin{figure}[t]
\centering
\begin{minipage}[c]{0.58\textwidth}
\centering
\captionof{table}{\textbf{Comparison showing gaze necessity on Qwen2.5-VL-7B.} Gaze improves performance over no-gaze and random-gaze.}
\label{tab:compare_necessity}
\vspace{6pt}
\setlength{\tabcolsep}{4.5pt} 
\resizebox{\columnwidth}{!}{
\begin{tabular}{l | c c  c c | c}
\toprule
  & \cellcolor[HTML]{E7F2DC}\textbf{OTP} & \cellcolor[HTML]{FBF5DC}\textbf{OI (Hard)} & 
\cellcolor[HTML]{FBF5DC}\textbf{FAP} & 
\cellcolor[HTML]{FBE4E7}\textbf{OAA} & \textbf{Avg.} \\ 
\midrule
 w/o Gaze & 0.432 & 0.479 & 0.457 & 0.250 & 0.405 \\
 w/ Gaze  & \textbf{0.471} & \textbf{0.530} & \textbf{0.540} & \textbf{0.276} & \textbf{0.454} \\
 \rowcolor{lightgray} \textbf{↑ (\%)} & +9.0\% & +10.6\% & +18.2\% & +10.4\% & +12.3\% \\
 \midrule
 w/ Random Gaze & 0.281 & 0.383 & 0.395 & 0.121 & 0.295 \\
\bottomrule
\end{tabular}
}
\end{minipage}
\hfill
\begin{minipage}[c]{0.38\textwidth}
\centering
\includegraphics[width=\columnwidth]{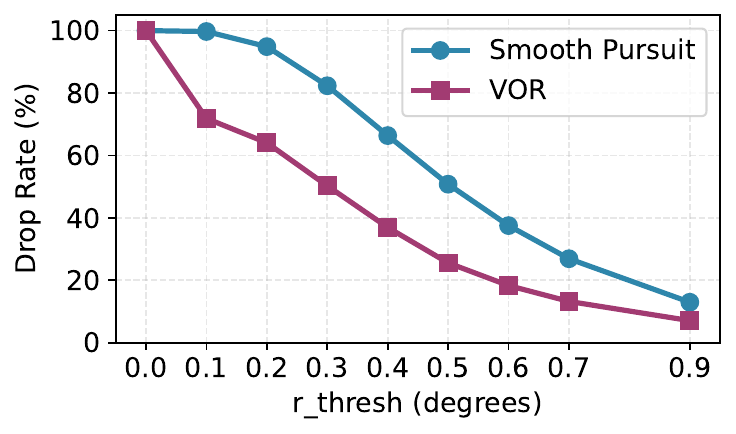}
\caption{\small\textbf{SP/VOR drop rate depending on fixation filtering.} 
}
\label{fig:drop_rate}
\end{minipage}
\end{figure}

Using HoloAssist IMU, head-pose, and camera parameters, we identify SP (Smooth Pursuit) and VOR (Vestibulo-Ocular Reflex) and evaluate their drop rates under fixation filtering for varying $r_{\text{thresh}}$ (\cref{fig:drop_rate}). 
For interpretability, $r_{\text{thresh}}$ is shown in visual degrees.
Using $r_{\text{thresh}}=0.3$, the drop rates are 82.3\% (SP) and 50.2\% (VOR).
We intentionally adopt a \textit{conservative} fixation definition to reduce temporal ambiguity, stabilize intervention timing, and enable fair and reproducible model comparison as a benchmark. 
Moreover, since most public egocentric video datasets do not provide IMU or head-pose signals, this gaze-only formulation makes our data generation pipeline more scalable and broadly applicable. 
We fully agree that incorporating dynamic interaction moments is an important direction for future work.

\subsection{Effect of Gaze Signals}
In \cref{tab:compare_necessity}, we present a counterfactual comparison across models w/ gaze, w/o gaze, and w/ random gaze.
Incorporating gaze signals consistently improves performance across all tasks compared to the no-gaze setting, with notable gains in OI (+10.6\%) and FAP (+18.2\%), leading to an overall improvement of +12.3\%. This indicates that gaze provides critical cues for identifying user attention and grounding intent over time. In contrast, replacing true gaze with random gaze significantly degrades performance across all metrics, even falling well below the no-gaze baseline. 
Overall, these results demonstrate that accurate gaze information is essential for reliable intent reasoning and cannot be substituted with noisy or uninformative proxies.

\subsection{Evaluation Under a Unified Frame Budget}
\label{app:unified_frames}

In our main experiments, we follow the official inference protocols of
each baseline, consistent with prior streaming benchmarks such as
ViSpeak~\citep{fu2025vispeak}.
To examine the effect of different frame sampling strategies,
we additionally evaluate open-source MLLMs under a unified
32-frame budget, as shown in Table~\ref{tab:unified_frames}.
The results reveal that increasing the number of input frames
does not necessarily improve performance.
For instance, VITA-1.5 experiences a substantial performance drop
from 0.384 to 0.175, suggesting sensitivity to deviations from
its default inference configuration.
Similarly, Qwen2.5-VL decreases from 0.478 to 0.422,
whereas InternVL3.5 improves from 0.444 to 0.494.
Meanwhile, EgoGPT and MiniCPM-V maintain identical performance,
as their original configurations already use 32 frames.
These results highlight the importance of model-specific inference
settings and motivate reporting both official-protocol results
and controlled frame-budget comparisons.

\begin{table*}[t]
    \centering
    \footnotesize
    \caption{
    \textbf{Comparison of open-source MLLMs under a unified 32-frame budget.}
    We evaluate all models using 32 input frames to control for
    differences in frame budgets.
    Performance may differ from the main results due to changes
    in model-specific inference configurations.
    The best performance in each task is \textbf{bolded}.
    }
    \resizebox{\linewidth}{!}{
    \begin{tabular}{lc|cccc|cccc|cc|c}
        \toprule
        \multirow{2}{*}{Method} &
        \multirow{2}{*}{Frames} &
        \multicolumn{4}{c|}{\cellcolor[HTML]{E7F2DC}\textbf{Past}} &
        \multicolumn{4}{c|}{\cellcolor[HTML]{FBF5DC}\textbf{Present}} &
        \multicolumn{2}{c|}{\cellcolor[HTML]{FBE4E7}\textbf{Proactive}} &
        \multirow{2}{*}{\textbf{Overall}} \\

        & &
        \cellcolor[HTML]{E7F2DC}NFI &
        \cellcolor[HTML]{E7F2DC}OTP &
        \cellcolor[HTML]{E7F2DC}SR &
        \cellcolor[HTML]{E7F2DC}GSM &
        \cellcolor[HTML]{FBF5DC}OI (E) &
        \cellcolor[HTML]{FBF5DC}OI (H) &
        \cellcolor[HTML]{FBF5DC}OAR &
        \cellcolor[HTML]{FBF5DC}FAP &
        \cellcolor[HTML]{FBE4E7}GTA &
        \cellcolor[HTML]{FBE4E7}OAA & \\

        \midrule

        InternVL3.5~\citep{wang2025internvl3} & 32 &
        0.511 & 0.407 & \textbf{0.578} & 0.511 &
        0.660 & 0.668 & 0.489 & 0.380 &
        \textbf{0.628} & 0.111 & \textbf{0.494} \\

        EgoGPT~\citep{yang2025egolife} & 32 &
        \textbf{0.521} & 0.393 & 0.453 & \textbf{0.550} &
        \textbf{0.690} & 0.620 & 0.304 & 0.368 &
        0.250 & 0.194 & 0.436 \\

        Qwen2.5-VL~\citep{wang2024qwen2vlenhancingvisionlanguagemodels} & 32 &
        0.169 & \textbf{0.445} & 0.246 & 0.538 &
        0.561 & \textbf{0.671} & \textbf{0.568} & \textbf{0.394} &
        0.463 & 0.163 & 0.422 \\

        MiniCPM-V~\citep{yao2024minicpm} & 32 &
        0.430 & 0.374 & 0.354 & 0.296 &
        0.334 & 0.379 & 0.438 & 0.345 &
        0.480 & \textbf{0.216} & 0.365 \\

        VITA 1.5~\citep{fu2025vita} & 32 &
        0.226 & 0.211 & 0.280 & 0.210 &
        0.275 & 0.149 & 0.040 & 0.250 &
        0.076 & 0.037 & 0.175 \\

        \bottomrule
    \end{tabular}
    }
    \label{tab:unified_frames}
\end{table*}

\subsection{Sensitivity to Proactive Evaluation Tolerance}
\label{app:proactive_tolerance}

Following prior proactive video understanding benchmarks
such as OVO-Bench~\citep{niu2025ovo},
we adopt an accuracy-based evaluation protocol for
consistent comparison across models.
To examine the sensitivity of this protocol to temporal
tolerance, we vary the tolerance window used to assign
positive labels ($\pm 10$, $\pm 15$, and $\pm 20$ seconds)
for our proactive tasks, Gaze-Triggered Alert (GTA)
and Object Appearance Anticipation (OAA).

As shown in \Cref{tab:proactive_tolerance},
the relative model rankings remain unchanged across
all three tolerance settings for both tasks.
Specifically, Qwen3-VL-8B consistently outperforms
TimeChat-Online and VITA-1.5, regardless of the
tolerance window.
Although absolute accuracy increases as the tolerance
window widens, the comparative ordering remains stable.
These results indicate that our conclusions about
relative model performance are robust to the choice
of temporal tolerance, rather than being driven by
a particular evaluation threshold.

\begin{table}[t]
\centering
\caption{
\textbf{Sensitivity to proactive evaluation tolerance.}
Accuracy under different temporal tolerance windows
for GTA and OAA. Model rankings are shown in parentheses.
}
\label{tab:proactive_tolerance}
\small
\setlength{\tabcolsep}{5pt}
\vspace{5pt}
\begin{tabular}{llccc}
\toprule
\textbf{Task} &
\textbf{Model} &
$\boldsymbol{\pm 10}$s &
$\boldsymbol{\pm 15}$s &
$\boldsymbol{\pm 20}$s \\
\midrule

\multirow{3}{*}{GTA}
& Qwen3-VL-8B~\citep{yang2025qwen3}
& 0.4733 (1)
& 0.4915 (1)
& 0.5096 (1) \\
& TimeChat-Online~\citep{yao2025timechat}
& 0.2448 (2)
& 0.2834 (2)
& 0.3227 (2) \\
& VITA-1.5~\citep{fu2025vita}
& 0.0564 (3)
& 0.0668 (3)
& 0.0784 (3) \\

\midrule

\multirow{3}{*}{OAA}
& Qwen3-VL-8B~\citep{yang2025qwen3}
& 0.6321 (1)
& 0.6352 (1)
& 0.6379 (1) \\
& TimeChat-Online~\citep{yao2025timechat}
& 0.1966 (2)
& 0.2264 (2)
& 0.2575 (2) \\
& VITA-1.5~\citep{fu2025vita}
& 0.0511 (3)
& 0.0585 (3)
& 0.0646 (3) \\

\bottomrule
\end{tabular}
\end{table}

\section{\ours{} Data Details}\label{sec:appendix-data-detail}

\subsection{Data Statistics}

\textbf{Overall statistics.} 
We provide \ours{} data statistics in \cref{fig:task_proportion,fig:data_stat,table:task_stats}.
From \cref{fig:task_proportion}:
(a) The videos cover three domains—cooking, laboratory work, and assembly—with cooking comprising the majority. This distribution reflects realistic egocentric environments while still offering diverse interaction patterns.
(b) The word cloud highlights frequently appearing FOV objects such as hands, knives, gloves, bottles, and cutting boards, indicating that the extracted objects largely correspond to action-related tools and manipulable items.
(c) Video lengths vary from just a few minutes to more than half an hour, with most clips between 5 and 15 minutes. This range provides both short segments for immediate reasoning and longer sequences suitable for multi-step or extended interactions.

\noindent\textbf{Attribute statistics of OAR task.}
For the OAR (Object Attribute Recognition) task, we generate attribute-focused questions centered on the gazed object’s properties. The distribution covers \textit{color} (65.61\%), \textit{material} (19.3\%), \textit{state} (5.5\%), \textit{texture} (5.3\%), and \textit{shape} (4.1\%), reflecting the dominant attribute types encountered in egocentric scenes.

\subsection{QA Examples}

\begin{figure*}[t]
    \centering
    {
    \includegraphics[width=0.9\textwidth]{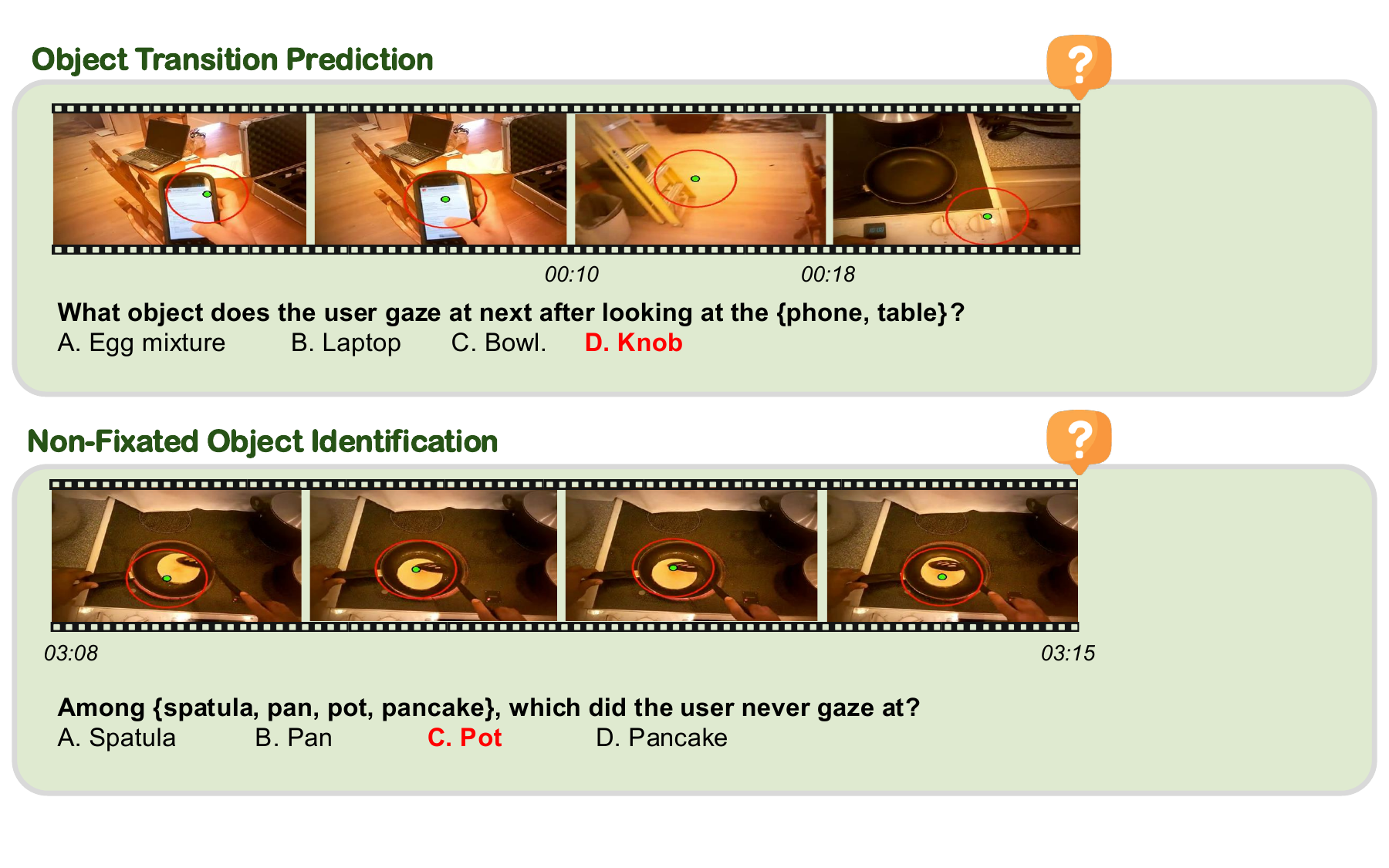}}
    \caption{
    \textbf{\ours{} data example for OTP and NFI task.}
    }
    \label{fig:task1}
    \vspace{-0.3in}
\end{figure*}

\begin{figure*}[t]
    \centering
    {
    \includegraphics[width=0.9\textwidth]{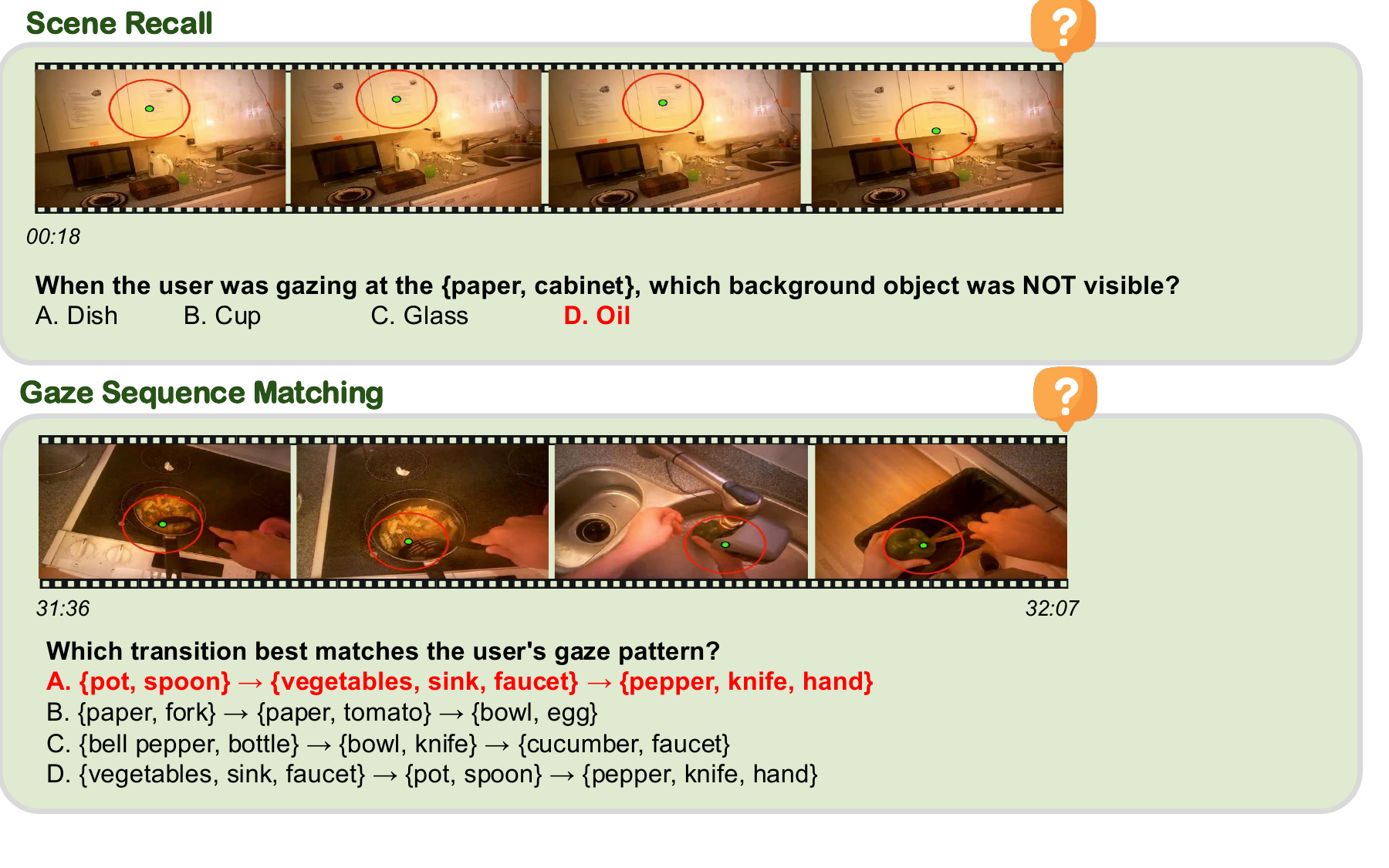}}
    \caption{
    \textbf{\ours{} data example for SR and GSM task.}
    }
    \label{fig:task2}
\end{figure*}

\begin{figure*}[t]
    \centering
    {
    \includegraphics[width=0.9\textwidth]{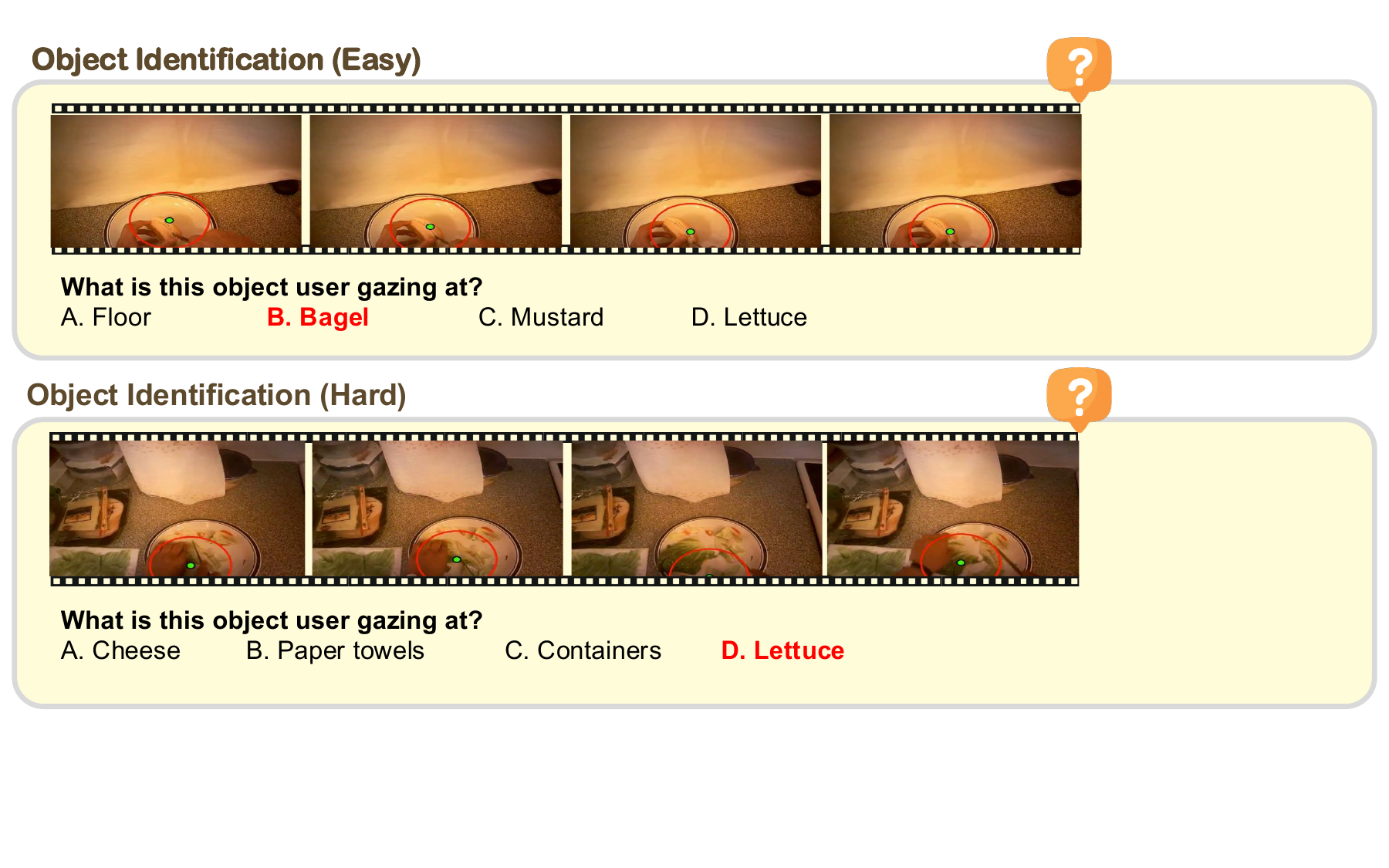}}
    \caption{
    \textbf{\ours{} data example for OI (Easy/Hard) task.}
    }
    \label{fig:task3}
    \vspace{-0.3in}
\end{figure*}

\begin{figure*}[t]
    \centering
    {
    \includegraphics[width=0.9\textwidth]{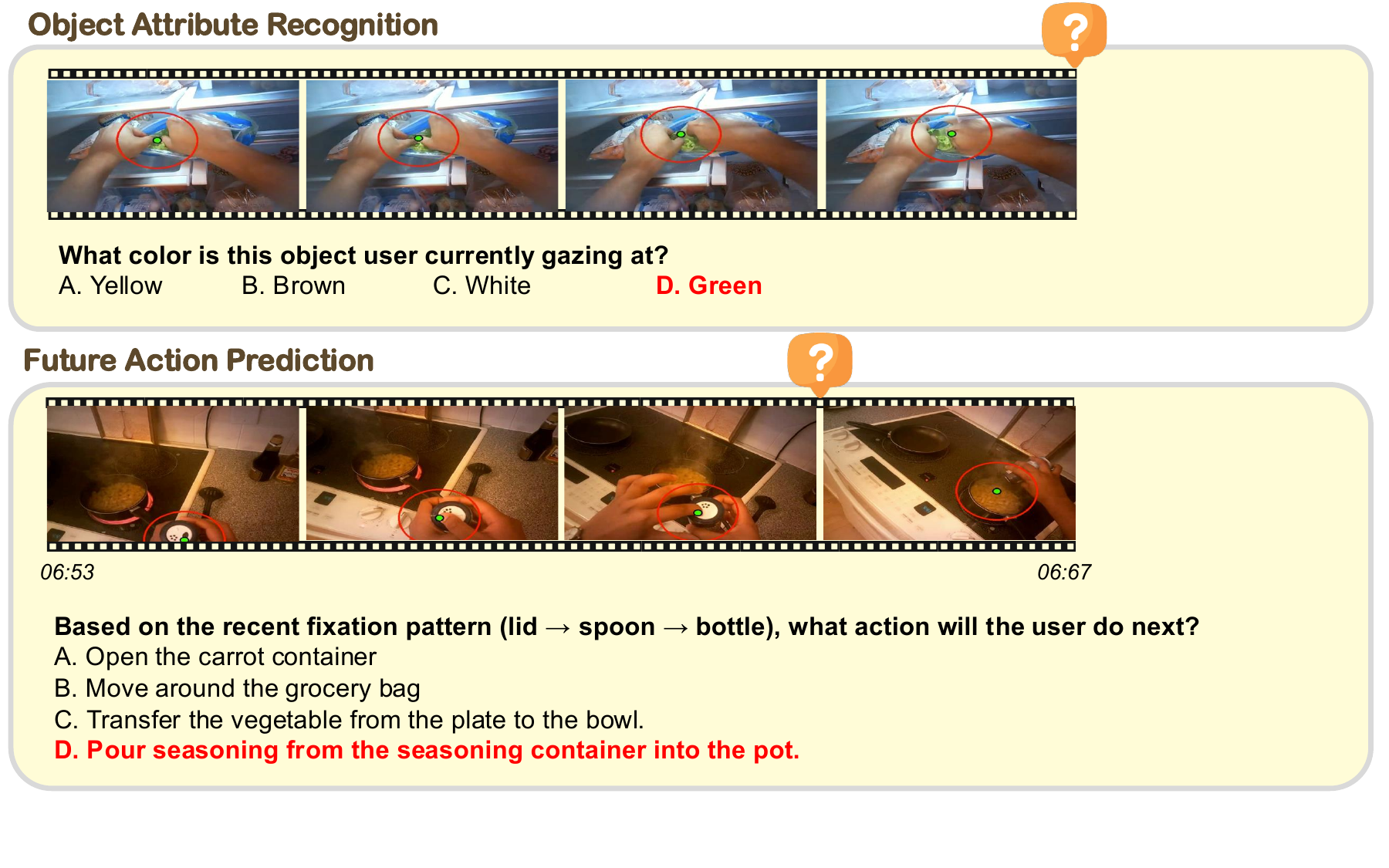}}
    \caption{
    \textbf{\ours{} data example for OAR and FAP task.}
    }
    \label{fig:task4}
\end{figure*}

\begin{figure*}[t]
    \centering
    {
    \includegraphics[width=0.9\textwidth]{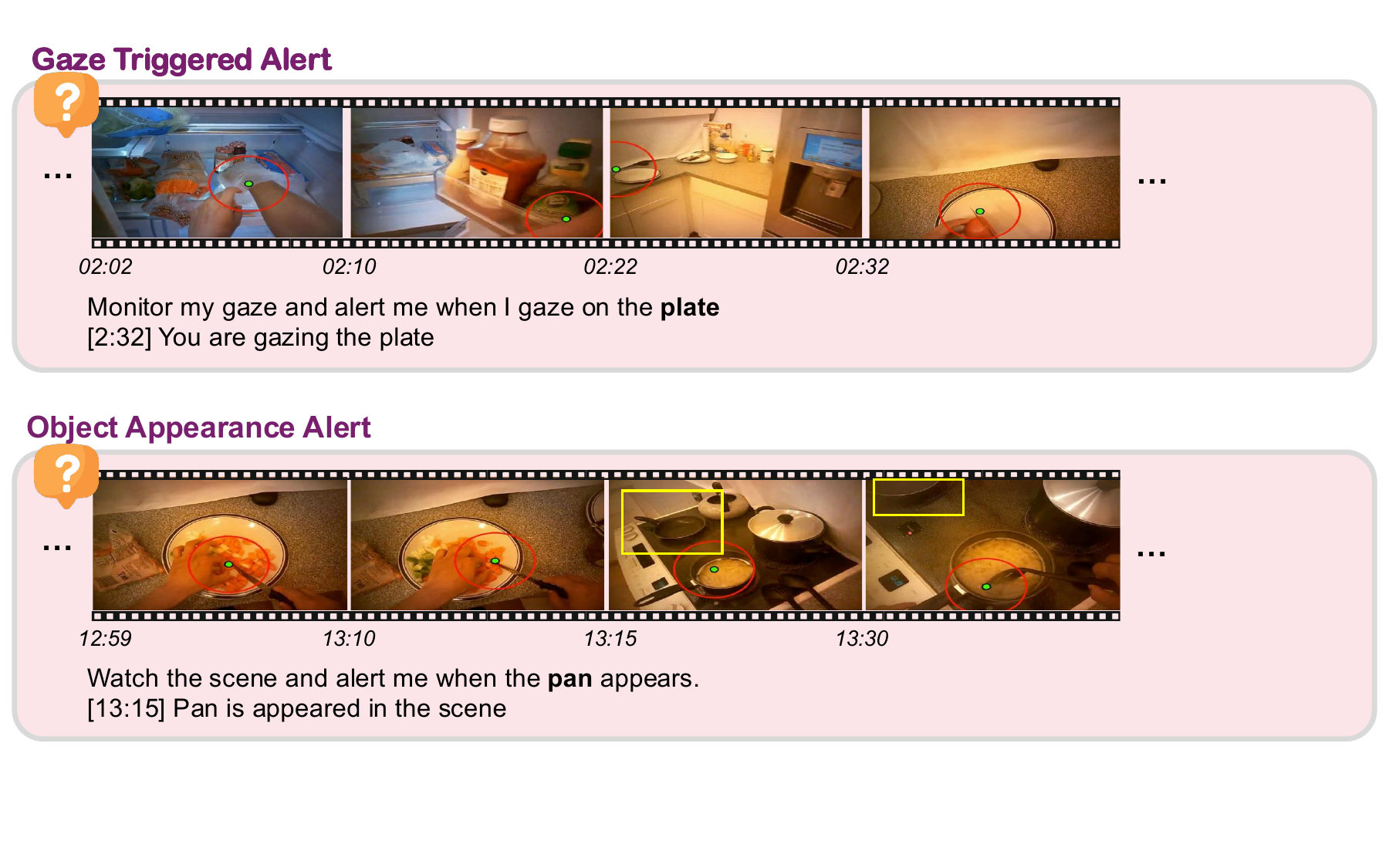}}
    \caption{
    \textbf{\ours{} data example for GTA and OAA task.}
    }
    \label{fig:task5}
\end{figure*}

We provide data examples per each \ours{} task for better understanding in \cref{fig:task1,fig:task2,fig:task3,fig:task4,fig:task5}. 
An orange question mark in each task figure denotes a query time.

\begin{table}[t]
\centering
\begin{minipage}[t]{0.50\textwidth}
\centering
\small
\caption{\textbf{Statistics of StreamGaze tasks.}}
\label{table:task_stats}
\vspace{4pt} 
\resizebox{\columnwidth}{!}{%
\begin{tabular}{l r}
\toprule
\textbf{Task} & \textbf{\# Samples} \\
\midrule
Gaze Sequence Matching & 186 \\
Non-Fixated Object Identification & 650 \\
Object Transition Prediction & 494 \\
Scene Recall & 211 \\
Future Action Prediction & 921 \\
Object Attribute Recognition & 1,419 \\
Object Identification (Easy) & 1,487 \\
Object Identification (Hard) & 1,005 \\
Gaze-Triggered Alert & 283 \\
Object Appearance Alert & 1,865 \\
\midrule
\textbf{Total} & \textbf{8,521} \\
\bottomrule
\end{tabular}%
}
\end{minipage}
\hfill
\begin{minipage}[t]{0.43\textwidth}
\centering
\small
\caption{\textbf{Licenses of datasets and models.}}
\label{table:license}
\vspace{4pt} 
\resizebox{\columnwidth}{!}{%
\begin{tabular}{c c}
\toprule
\textbf{Resource} & \textbf{License} \\
\midrule
EgoExoLearn & MIT License (\href{https://github.com/OpenGVLab/EgoExoLearn/blob/main/LICENSE}{link}) \\
HoloAssist & CDLAv2 License (\href{https://cdla.dev/permissive-2-0/}{link}) \\
ViSpeak & Apache-2.0 (\href{https://huggingface.co/fushh7/ViSpeak-s3}{link}) \\
InternVL 3.5 & Apache-2.0 (\href{https://huggingface.co/OpenGVLab/InternVL3_5-3B8-HF}{link}) \\
Qwen3 & Apache-2.0 (\href{https://huggingface.co/Qwen/Qwen3-235B-A22B-Thinking-2507-FP8/blob/main/LICENSE}{link}) \\
\bottomrule
\end{tabular}%
}
\end{minipage}
\end{table}

\section{License}\label{sec:appendix-license}
We provide licenses of datasets and models used in \cref{table:license}.

\input{table/object_prompts}
\input{table/gaze_reasoning_prompt}

\begin{figure*}[t]
    \centering
    {
    \includegraphics[width=0.8\textwidth]{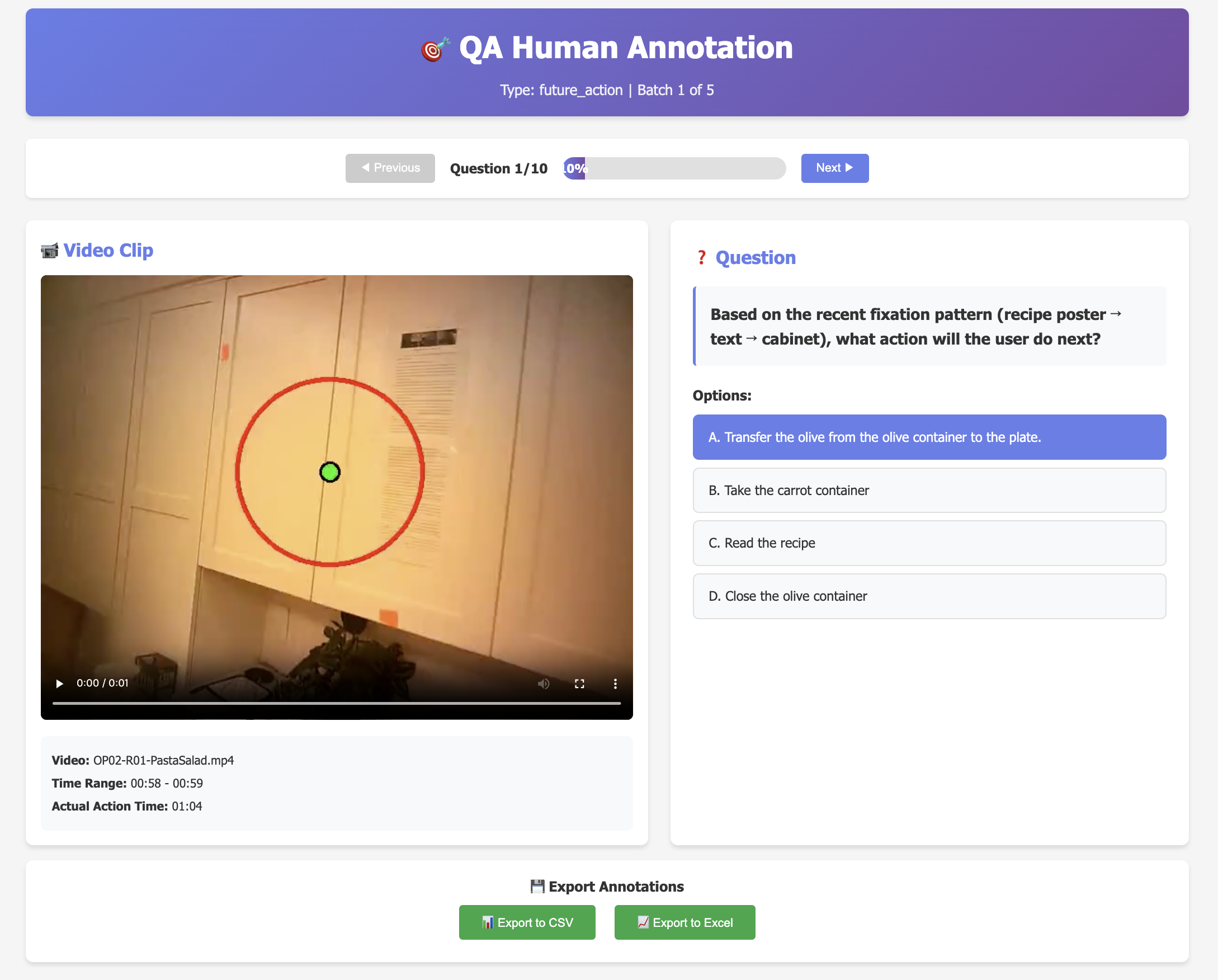}}
    \caption{
    \textbf{HTML for human oracle evaluation of \ours{}.}
    }
    \label{fig:qa_html}
\end{figure*}

%% file: table/algo_pipeline.tex
\begin{algorithm}[t]
\caption{\textbf{Gaze-Guided Data Construction Pipeline}}
\label{alg:pipeline}
\begin{algorithmic}[1]

\State \textbf{Input:} $\mathcal{V}=\{I_t\}_{1:T}$, $\mathcal{G} = \{(x_t, y_t)\}_{t=1}^{T}$, $r_{\text{thresh}}, \tau_{\text{dur}}, \tau_{\text{scene}}$
\Statex

\State \texttt{\# Fixation Extraction}
\State $\mathcal{F} \gets \emptyset$
\For{each candidate interval $[t_i^s, t_i^e]$}
    \State Compute centroid $(\bar{x}_i, \bar{y}_i)$
    \If{$\forall\, t \in [t_i^s, t_i^e],\ 
        \|(x_t, y_t) - (\bar{x}_i, \bar{y}_i)\|_2 \le r_{\text{thresh}}$}
        \If{$t_i^e - t_i^s \ge \tau_{\text{dur}}$}
            \State $S_{\min} \gets 
                \min_{t = t_i^s}^{t_i^e - 1} \rho(H_t, H_{t+1})$
            \If{$S_{\min} \ge \tau_{\text{scene}}$}
                \State $\mathcal{F} \gets \mathcal{F} \cup 
                    \{(\bar{x}_i, \bar{y}_i, t_i^s, t_i^e)\}$
            \EndIf
        \EndIf
    \EndIf
\EndFor

\Statex
\State \texttt{\# Object and scanpath extraction}
\For{each $f_i\in\mathcal{F}$}
    \For{$t = t_i^s \dots t_i^e$}
        \State Define $\mathcal{R}_{i,t}^{\text{fov}}$ and $\mathcal{R}_{i,t}^{\text{out}}$
    \EndFor
    \State $\mathcal{O}_i^{\text{fov}} \gets \mathrm{MLLM}(\mathcal{R}_i^{\text{fov}})$
    \State $\mathcal{O}_i^{\text{out}} \gets \mathrm{MLLM}(\mathcal{R}_i^{\text{out}})$
\EndFor

\State $\mathcal{S} \gets \{(\mathcal{O}_i^{\text{fov}},\mathcal{O}_i^{\text{out}})\}_{i=1}^{N}$

\Statex
\State \texttt{\# QA Generation}

\end{algorithmic}
\end{algorithm}

%% file: table/sft.tex
\begin{table*}[t]
  \centering
  \footnotesize
\caption{\textbf{Fine-tuning results on ViSpeak~\cite{fu2025vispeak}.}
We construct four variants of training data for our fine-tuning experiments, drawing from both external resources~\cite{hdepic,peng2025eye} and our own automatically generated data based on Ego4D-Gaze~\cite{grauman2022ego4d} and EgoExoLearn~\cite{huang2024egoexolearn}.  
}
\label{tab:sft_result}
  \resizebox{\linewidth}{!}{
  \begin{tabular}{ccc|cccc|cccc|cc|c}
    \toprule
      & 
    \multicolumn{1}{c}{\textit{Fine-tuning datasets}} & 
     & 
    \multicolumn{4}{c|}{\cellcolor[HTML]{E7F2DC}\textbf{Past}} & 
    \multicolumn{4}{c|}{\cellcolor[HTML]{FBF5DC}\textbf{Present}} & 
    \multicolumn{2}{c|}{\cellcolor[HTML]{FBE4E7}\textbf{Proactive}} & 
    \multirow{2}{*}{\textbf{Overall}} \\

    External & EgoExoLearn & Ego4D-Gaze & 
    \cellcolor[HTML]{E7F2DC}NFI & 
    \cellcolor[HTML]{E7F2DC}OTP & 
    \cellcolor[HTML]{E7F2DC}SR & 
    \cellcolor[HTML]{E7F2DC}GSM & 
    \cellcolor[HTML]{FBF5DC}OI (E) & 
    \cellcolor[HTML]{FBF5DC}OI (H) & 
    \cellcolor[HTML]{FBF5DC}OAR & 
    \cellcolor[HTML]{FBF5DC}FAP & 
    \cellcolor[HTML]{FBE4E7}GTA & 
    \cellcolor[HTML]{FBE4E7}OAA & 
    \\

    \midrule
     \rowcolor{lightgray} &  &  &  0.526 & 0.486 & 0.341 & 0.418 &
    0.635 & 0.477 & 0.413 & 0.489 &
    0.504 & 0.502 &
    0.479 \\
    \ding{51} &  &  & 0.467
    & 0.477 & 0.300 & 0.347 & 0.368 
    & 0.562 & 0.420 & 0.392 & 0.147
    & 0.381 & 0.386 \\

    \ding{51} &\ding{51} &  & 0.592
    & 0.563 & 0.469 & 0.385 & 0.622 
    & 0.616 & 0.544 & 0.359 & 0.474
    & 0.792 & 0.542 \\

    \ding{51} &  & \ding{51} & 0.517
    & 0.545 & 0.495 & 0.373 & 0.540 
    & 0.452 & 0.496 & 0.351 & 0.487 
    & 0.737 & 0.500 \\

    \ding{51} & \ding{51} & \ding{51} & 0.583
    & 0.540 & 0.495 & 0.388 & 0.601 
    & 0.641 & 0.575 & 0.331 & 0.409 
    & 0.651 & 0.521 \\
    
    \bottomrule
  \end{tabular}
  }
\end{table*}

%% file: table/prompt_ablation_total.tex
\begin{table*}[t]
  \centering
  \footnotesize
  \caption{\textbf{Detailed performance of gaze input prompting across each \ours{} task.}
  We ablate effect of gaze input prompting on GPT-4o~\cite{openai2024gpt4technicalreport}, InternVL3.5~\cite{wang2025internvl3} and ViSpeak~\cite{fu2025vispeak}.
  }
  \resizebox{\linewidth}{!}{
  \begin{tabular}{lcc|cccc|cccc|cc|c}
    \toprule
    \multirow{2}{*}{Method} & 
    \multirow{2}{*}{Params} & 
    \multirow{2}{*}{Frames} & 
    \multicolumn{4}{c|}{\cellcolor[HTML]{E7F2DC}\textbf{Past}} & 
    \multicolumn{4}{c|}{\cellcolor[HTML]{FBF5DC}\textbf{Present}} & 
    \multicolumn{2}{c|}{\cellcolor[HTML]{FBE4E7}\textbf{Proactive}} & 
    \multirow{2}{*}{\textbf{Overall}} \\

    & & & 
    \cellcolor[HTML]{E7F2DC}NFI & 
    \cellcolor[HTML]{E7F2DC}OTP & 
    \cellcolor[HTML]{E7F2DC}SR & 
    \cellcolor[HTML]{E7F2DC}GSM & 
    \cellcolor[HTML]{FBF5DC}OI (E) & 
    \cellcolor[HTML]{FBF5DC}OI (H) & 
    \cellcolor[HTML]{FBF5DC}OAR & 
    \cellcolor[HTML]{FBF5DC}FAP & 
    \cellcolor[HTML]{FBE4E7}GTA & 
    \cellcolor[HTML]{FBE4E7}OAA & 
    \\

    \midrule
    GPT-4o~\cite{openai2024gpt4technicalreport} & - & 16 & 
    0.601 & 0.459 & 0.507 & 0.607 & 
    0.725 & 0.731 & 0.594 & 0.375 & 
    0.608 & 0.148 & 0.536 \\
    ~+ visual prompt & - & 16 & 
    0.601 & 0.449 & 0.535 & 0.580 & 
    0.729 & 0.730 & 0.596 & 0.370 & 
    0.597 & 0.149 & 0.535 \\
    
    \midrule
    
    InternVL3.5~\cite{wang2025internvl3} & 8B & Adaptive & 
    0.469 & 0.370 & 0.526 & 0.510 & 
    0.626 & 0.637 & 0.463 & 0.382 &
    0.376 & 0.048 & 0.441 \\
    ~+ visual prompt & 8B & Adaptive & 
    0.490 & 0.311 & 0.573 & 0.548 & 
    0.627 & 0.628 & 0.466 & 0.372 &
    0.373 & 0.051 & 0.444 \\
    \midrule
        
    ViSpeak~\cite{fu2025vispeak} & 7B & 1 fps & 
    0.493 & 0.374 & 0.417 & 0.521 &
    0.591 & 0.560 & 0.438 & 0.336 &
    0.625 & 0.334 &
    0.469 \\
    ~+ visual prompt & 7B & 1 fps & 
    0.463 & 0.358 & 0.417 & 0.473 &
    0.572 & 0.581 & 0.406 & 0.309 &
    0.635 & 0.458 &
    0.467 \\
    
    \bottomrule

  \end{tabular}
  }
  \label{tab:prompt_ablation_total}
\end{table*}

%% file: table/object_prompts.tex
\begin{figure*}[t]
\centering
\begin{tcolorbox}[
    colback=gray!10,
    colframe=black!80,
    title={Prompt for FOV region object extraction\label{box:obj-extraction-fov}},
    width=\textwidth,
    boxrule=0.8pt
]
\small
\begin{verbatim}
Context: {action_caption}

Known objects: {object_pool}
IMPORTANT: Reuse these exact names if you see the same objects. 
Only use new names for clearly different objects.

Analyze this video clip and return a JSON with this exact structure:

{
  "scene_caption": "Brief 2-3 sentence description of the overall scene 
  and what's happening",
  "gaze_object": {
    "object_identity": "name of object at coordinate (x, y)",
    "detailed_caption": "Natural 3-5 sentence description of this object's 
    appearance and characteristics"
  },
  "other_objects": [
    {
      "object_identity": "object_name (e.g., 'cup', 'person wearing blue shirt')",
      "detailed_caption": "Two sentences describing the object's appearance, 
      attributes, and spatial position relative to other objects"
    }
  ]
}

Person rules: Only include if full/upper body visible (not just hands/arms). 
CRITICAL: object_identity must be "person wearing [clothing description]" 
NOT just "person".

Return only valid JSON. Be concise and direct.
\end{verbatim}
\end{tcolorbox}
\caption{\textbf{Prompt for FOV region object extraction.}}
\label{fig:prompt-fov}
\end{figure*}

\begin{figure*}[t]
\centering
\begin{tcolorbox}[
    colback=gray!10,
    colframe=black!80,
    title={Prompt for out-of-FOV region object extraction\label{box:obj-extraction-fov-out}},
    width=\textwidth,
    boxrule=0.8pt
]
\small
\begin{verbatim}

Context: {action_caption}

Known objects: {object_pool}
IMPORTANT: Reuse these exact names if you see the same objects. 
Only use new names for clearly different objects.

Analyze this video clip (with the center region masked) and return a JSON 
with this structure:

{
  "other_objects": [
    {
      "object_identity": "object_name (e.g., 'bottle', 'person wearing 
      green jacket')",
      "detailed_caption": "Two sentences describing the object's appearance, 
      attributes, and spatial position"
    }
  ]
}

Focus only on objects OUTSIDE the black masked region.

Person rules: Only include if full/upper body visible (not just hands/arms). 
CRITICAL: object_identity must be "person wearing [clothing description]" 
NOT just "person".

Return only valid JSON.

\end{verbatim}
\end{tcolorbox}
\caption{\textbf{Prompt for out-of-FOV region object extraction.}}
\label{fig:prompt-out-fov}
\end{figure*}

%% file: table/gaze_reasoning_prompt.tex
\begin{figure*}[t]
\centering
\begin{tcolorbox}[
    colback=gray!10,
    colframe=black!80,
    title={Prompt for text-based reasoning
    },
    width=\textwidth,
    boxrule=0.8pt
]
\small
\begin{verbatim}
You are an expert at gaze-conditioned streaming video reasoning. 

Rules: 
1) Only use visual evidence up to the given timestamp. 
2) Do your step-by-step reasoning ONLY inside <think>...</think>. 
3) Give exactly ONE final choice (A/B/C/D) or yes/no with its short text 
inside <answer>...</answer>. 

{Question}

\end{verbatim}
\end{tcolorbox}
\caption{\textbf{Prompt for text-based reasoning.}}
\label{fig:text-reasoning-input}
\end{figure*}

\begin{figure*}[t]
\centering
\begin{tcolorbox}[
    colback=gray!10,
    colframe=black!80,
    title={Prompt for gaze-based reasoning
    },
    width=\textwidth,
    boxrule=0.8pt
]
\small
\begin{verbatim}
You are an expert at gaze-conditioned streaming video reasoning. 

Rules: 
1) Only use visual evidence up to the given timestamp (no future leakage). 
2) Do your step-by-step reasoning ONLY inside <think>...</think>. 
3) Return exactly three tags in order: <gaze>...</gaze> <think>...</think> 
<answer>...</answer>. 
4) In <gaze>, estimate the most recent gaze center and FOV radius from 
the green dot and red circle in the frames up to the timestamp. 
5) If the green dot/Red circle is not visible, output "unknown" in <gaze> 
but still reason and answer using other available evidence. 
6) In <answer>, give exactly ONE final choice (A/B/C/D) or Yes/No 
with a short text. 

{Question}
\end{verbatim}
\end{tcolorbox}
\caption{\textbf{Prompt for gaze-based reasoning.}}
\label{fig:gaze-reasoning-input}
\end{figure*}

\begin{figure*}[t]
\centering
\begin{tcolorbox}[
    colback=gray!10,
    colframe=black!80,
    title={Prompt for visual-based reasoning
    },
    width=\textwidth,
    boxrule=0.8pt
]
\small
\begin{verbatim}
You are an expert at gaze-conditioned streaming video reasoning. 

Rules: 

1) Only use visual evidence up to the given timestamp (no future leakage). 
2) Produce your reasoning ONLY inside <think>...</think> and never reveal 
reasoning outside the tag. 
3) Return exactly FOUR tags in this order: <gaze>...</gaze> <objects>...
</objects> <think>...</think> <answer>...</answer>. 
4) In <gaze>, estimate the most recent gaze center and FOV radius from 
the green dot and red circle, or output "unknown" if not visible. 
5) In <objects>, ground the most relevant objects inside or closest to the 
gaze/FOV region; for each object provide: object_name, bbox_x1, bbox_y1, 
bbox_x2, bbox_y2 normalized to [0,1], or output "none" if no relevant 
object appears. 
6) In <think>, use the gaze and grounded objects to perform step-by-step 
reasoning and explain exclusion of irrelevant options. 
7) In <answer>, provide exactly ONE final label (A/B/C/D or Yes/No) 
with a short justification. 

{Question}
\end{verbatim}
\end{tcolorbox}
\caption{\textbf{Prompt for visual-based reasoning.}}
\label{fig:visual-reasoning-input}
\end{figure*}